\documentclass[letterpaper]{article} 
\usepackage[preprint]{aaai2027}  
\usepackage[hyphens]{url}  
\usepackage{graphicx} 
\usepackage{natbib}  
\usepackage{caption} 
\usepackage{algorithm}
\usepackage{algorithmic}

\usepackage{newfloat}
\usepackage{listings}
\DeclareCaptionStyle{ruled}{labelfont=normalfont,labelsep=colon,strut=off} 
\floatstyle{ruled}
\newfloat{listing}{tb}{lst}{}
\floatname{listing}{Listing}

\usepackage{booktabs}
\usepackage{colortbl}
\usepackage{amsmath}
\usepackage{amssymb}

\title{ManeuverFlow: Maneuver Hypothesis Generation with Future Evidence for Trajectory Evaluation and Refinement in End-to-End Autonomous Driving}
\author{
    Yuhao Wei\textsuperscript{\rm 1}, Yong Chen\textsuperscript{\rm 2}, Chen Yang\textsuperscript{\rm 1}, Yingchen Wang\textsuperscript{\rm 1}, Delin Ouyang\textsuperscript{\rm 1},\\
    Lingfeng Qi\textsuperscript{\rm 1}, Qi Lan\textsuperscript{\rm 1}, Zhanhua Zhang\textsuperscript{\rm 2}, Jie Li\textsuperscript{\rm 1}, Guofa Li\textsuperscript{\rm 1,*}
}
\affiliations{
    \textsuperscript{\rm 1}Chongqing University\\
    \textsuperscript{\rm 2}Artificial Intelligence Center, Geely Automotive Research Institute
}

\usepackage{array}
\usepackage{multirow}
\makeatletter
\let\MFmainaddtocurcol\@addtocurcol
\makeatother
\usepackage{flafter}
\makeatletter
\let\MFappendixaddtocurcol\@addtocurcol
\let\@addtocurcol\MFmainaddtocurcol
\makeatother

\begin{document}

\maketitle
\begingroup
\renewcommand{\thefootnote}{\fnsymbol{footnote}}
\footnotetext[1]{\raggedright Corresponding author: \mbox{Guofa Li (liguofa@cqu.edu.cn)}.}
\endgroup

\begin{abstract}
End-to-end planning in complex scenes must model diverse driving intentions and select the most appropriate trajectory using scene context and future risks.
However, multi-query regression may collapse to one mode, while anchor-based and generative planners share a scaling bottleneck: small candidate sets limit maneuver coverage or discovery of high-quality trajectories, whereas large sets increase generation and scoring costs.
We propose \textbf{ManeuverFlow} to generate trajectories for distinct intentions in one step and evaluate and refine them using current and future evidence.
The model selects relevant maneuver modes, draws corresponding noise from the Gaussian mixture prior, and maps it to intention-specific trajectories through a maneuver-routed mixture-of-experts MeanFlow field.
Current features and future features predicted by the world model are aggregated along each candidate path into trajectory-specific evidence for PDM-factor prediction.
Using the same evidence, the refiner predicts local Frenet residuals for selected candidates; refined and retained candidates are then jointly re-scored for selection.
With a lightweight TransFuser-style ResNet-34 backbone, ManeuverFlow achieves 90.6 PDMS and 90.1 EPDMS on NAVSIM v1/v2 \texttt{navtest}, respectively.
On WorldEngine's recent 3DGS closed-loop benchmark, it attains the highest listed PDMS$^\ast$ scores of 75.19 and 74.58 under logged and IDM-controlled traffic, respectively.
Code: \pdfstartlink attr{/Border [0 0 0]} user{/Subtype /Link /A << /S /URI /URI (https://github.com/wyh-autonomy/maneuverflow) >>}\url{https://github.com/wyh-autonomy/maneuverflow}\pdfendlink.
\end{abstract}

\section{Introduction}

\begin{figure}[!t]
    \centering
    \includegraphics[width=\columnwidth]{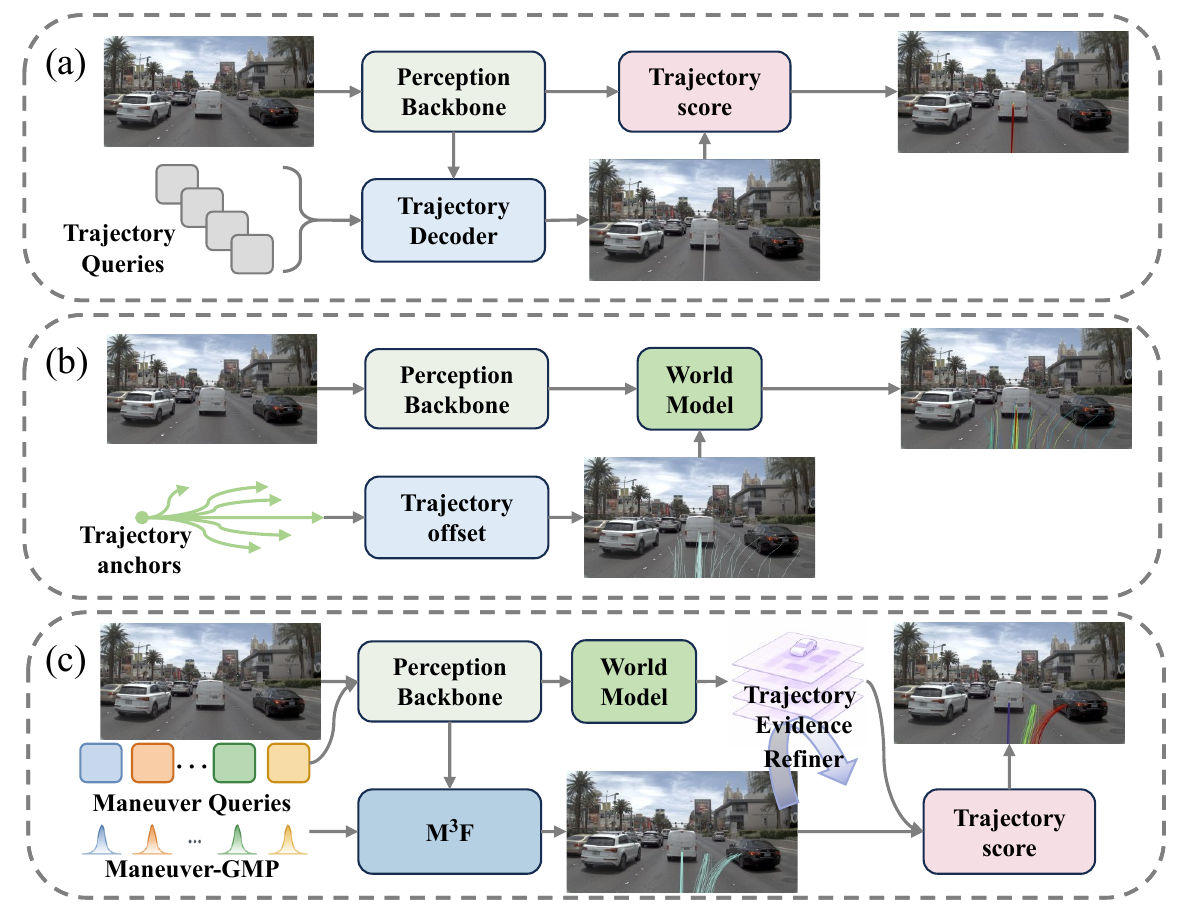}
    \caption{Candidate-based planning paradigms. (a) Query candidates without explicit mode structure, scored from current features. (b) Fixed-anchor candidates evaluated with per-candidate future rollouts. (c) ManeuverFlow with maneuver-structured generation, shared-future evidence, and Frenet refinement before joint re-scoring.}
    \label{fig:intro_comparison}
\end{figure}

End-to-end autonomous driving maps sensor observations, ego states, and navigation cues directly to future ego motion, unifying perception and planning under a single objective~\citep{chitta2022transfuser,hu2022stp3,hu2023planning,jiang2023vad,jia2023thinktwice,sun2025sparsedrive}. Complex traffic scenes often admit multiple plausible driving hypotheses, so a single trajectory cannot capture the underlying decision space. Multimodal motion forecasting represents this ambiguity with fixed anchor vocabularies or mode-specific queries~\citep{chai2020multipath,shi2022mtr}; recent planners further use learned queries, trajectory anchors, and diffusion- or flow-based generators to produce candidates, then select a final plan through candidate--scene interactions~\citep{jiang2026vadv2,li2024hydramdp,liao2025diffusiondrive,xing2025goalflow,wang2026meanfuser}. As shown in Fig.~\ref{fig:intro_comparison}(a), learned queries without explicit mode-level structure may concentrate in similar regions of trajectory space, producing redundant candidates and risking mode collapse.

\begin{figure}[!t]
    \centering
\includegraphics[width=\columnwidth]{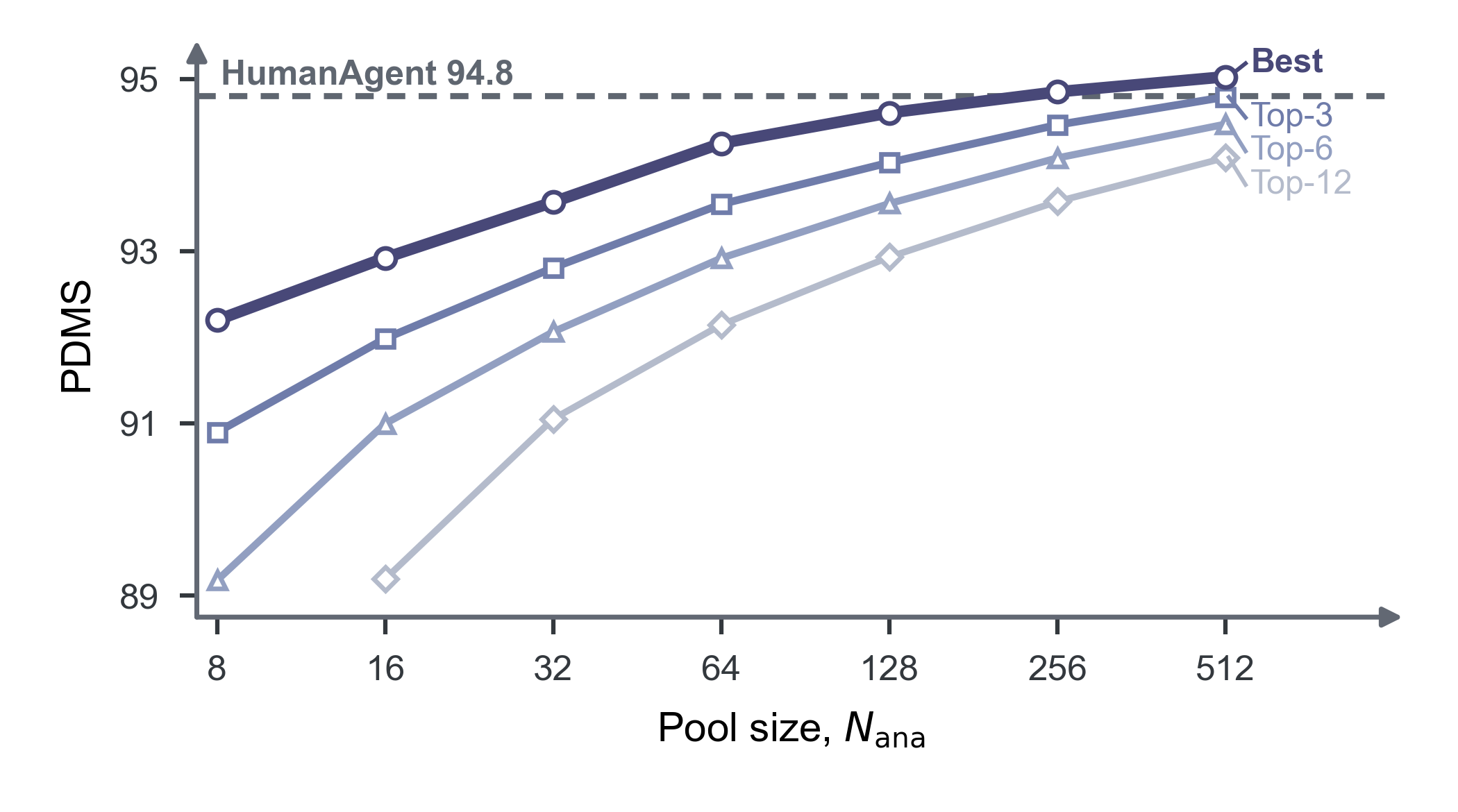}
    \caption{Post-hoc candidate-quality analysis of the frozen Maneuver-Mixture \mbox{MeanFlow} Field (M$^3$F) across 12,146 NAVSIM v1 \texttt{navtest} scenes, excluded from training and model selection. Curves report best and Top-$\kappa$ mean PDMS over nested pools; \mbox{HumanAgent} 94.8 is a dashed reference.}
    \label{fig:sampling_headroom}
\end{figure}

Standard diffusion and ODE-based flow sampling is iterative~\citep{lipman2023flowmatching,liu2023rectifiedflow}, although recent driving planners reduce the process to two steps in DiffusionDrive~\citep{liao2025diffusiondrive} or one step in GoalFlow and MeanFuser~\citep{xing2025goalflow,wang2026meanfuser}. MeanFuser generates multiple proposals in one MeanFlow step, but its Adaptive Reconstruction Module (ARM) attends to and fuses them into a single output trajectory; the initial hypotheses thus do not remain explicit candidates for downstream evaluation, refinement, and selection~\citep{wang2026meanfuser}. Fixed anchors encode predefined modes but may allocate part of the candidate budget to modes weakly relevant to the current scene. For a frozen generator, Fig.~\ref{fig:sampling_headroom} retrospectively shows that larger pools expose stronger candidates. This motivates improving candidate quality under a compact proposal budget through scene-adaptive hypothesis generation and evidence-conditioned local trajectory refinement.

Candidate generation also requires reliable selection. Candidate-tree and generation--evaluation planners pair ego hypotheses with conditional predictions or learned evaluators~\citep{huang2024dtpp,huang2025gendrive}, whereas current-only scorers do not explicitly query predicted future scene states along candidate paths. WoTE provides such evidence through one future rollout per candidate (Fig.~\ref{fig:intro_comparison}(b)), repeating future BEV prediction across candidate trajectories and multiple future time steps~\citep{li2025wote}.

Together, these limitations define a fixed-budget hypothesis-lifecycle problem. We therefore ask: \textit{How can a planner generate scene-relevant, mode-distinct hypotheses under a fixed budget and use current/future evidence to evaluate, refine, and select them?}

\begin{figure*}[!t]
    \centering
    \includegraphics[width=\textwidth]{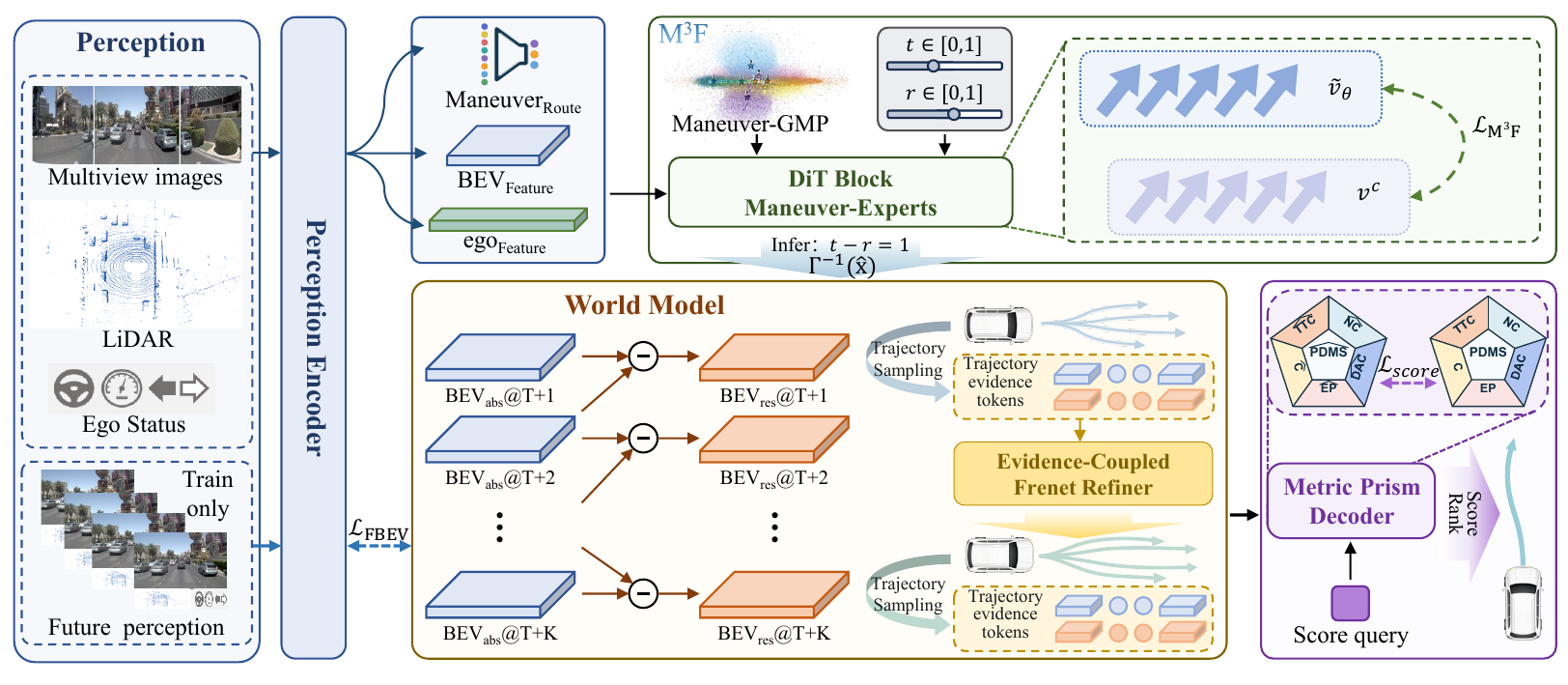}
    \caption{Overview of ManeuverFlow. Scene-conditioned M-GMP and the shared planning context drive one-step M$^3$F generation of base hypotheses, which Metric Prism pre-scores using trajectory-specific evidence from a shared future BEV field. ECFR refines the selected $R$ hypotheses, which join the retained bases for evidence re-query and joint re-scoring.}
    \label{fig:method_overview}
\end{figure*}

To this end, we propose \textbf{ManeuverFlow}, a unified \emph{Hypothesize--Verify--Improve} framework that preserves explicit maneuver-indexed hypotheses, as shown in Figs.~\ref{fig:intro_comparison}(c) and~\ref{fig:method_overview}. The Maneuver Gaussian Mixture Prior (M-GMP) predicts scene-conditioned weights over motion-defined components, and M$^3$F generates base hypotheses by maneuver-routed one-step transport. Once per planning step, the world model predicts one candidate-independent future BEV field; a learned trajectory-guided probe projects each hypothesis into the current and predicted fields to form path-specific Trajectory Evidence Memory. Metric Prism pre-scores candidates through learned PDM factors. The Evidence-Coupled Frenet Refiner (ECFR) uses the same evidence to predict local Frenet residuals for selected hypotheses. Refined hypotheses replace selected bases, re-query the shared field, and compete with retained hypotheses under the same Metric Prism scoring function.

Our main contributions are summarized as follows:
\begin{itemize}
    \item \textbf{Scene-adaptive maneuver generation.} We introduce a maneuver-structured generator that combines M-GMP with the maneuver-routed M$^3$F to model motion-defined maneuver components and generate scene-adaptive, intent-conditioned trajectory hypotheses as explicit alternatives in one step.
    \item \textbf{Future-evidence verification and refinement.} A learned probe projects each candidate onto a shared future BEV field to extract trajectory-specific evidence for PDM-aligned scoring and ECFR's local Frenet refinement. Metric Prism jointly re-scores refined and retained hypotheses for final selection.
    \item With a lightweight ResNet-34 TransFuser (TF) perception backbone, ManeuverFlow-TF achieves 90.58 PDMS and 90.14 EPDMS on NAVSIM v1/v2 \texttt{navtest}, respectively. Its camera-only Latent TransFuser (LTF) variant further obtains 75.19 and 74.58 PDMS$^\ast$ under replayed and reactive traffic on WorldEngine, delivering the highest listed composite planning scores across all reported benchmark and traffic settings.
\end{itemize}

\section{Related Work}

\subsection{End-to-End and Generative Planning}

End-to-end planners predict ego motion from sensor observations, ego states, and navigation commands. TransFuser fuses camera and LiDAR features, while later systems introduce planning-oriented queries and sparse scene representations~\citep{chitta2022transfuser,hu2023planning,jiang2023vad,sun2025sparsedrive}. Generative planners model multiple feasible decisions through structured latent spaces, diffusion, or flow matching: GenAD learns a trajectory latent space, Diffusion Planner and DiffusionDrive denoise trajectory distributions, and GoalFlow applies goal-conditioned flow matching~\citep{zheng2024genad,zheng2025diffusionplanner,liao2025diffusiondrive,xing2025goalflow}. MeanFuser combines a Gaussian-mixture source with one-step MeanFlow and aggregates sampled proposals into a single ARM output~\citep{geng2025meanflows,wang2026meanfuser}. MM-FM jointly designs a Gaussian-mixture source and mode-dependent coupling for general multimodal generation~\citep{luo2026multimodalflow}. A challenge is to organize driving-specific motion modes and retain diverse hypotheses as an explicit interface for evaluation and refinement.

\subsection{World Models for Planning}

Driving world models predict visual observations, occupancy, or latent states to support planning. MILE jointly learns latent dynamics and a driving policy, Think2Drive trains a planner through interaction with a compact latent world model, and LAW learns self-supervised latent dynamics~\citep{hu2022mile,li2024think2drive,li2025law}. Drive-WM, Drive-OccWorld, World4Drive, and WorldRFT use predicted visual, occupancy, or latent futures for planning evaluation and optimization~\citep{wang2023drivewm,yang2025driveoccworld,zheng2025world4drive,yang2025worldrft}. For online evaluation, WoTE runs a recurrent BEV world model for each anchor-refined candidate and selects the highest-scoring trajectory~\citep{li2025wote}. Candidate-conditioned rollouts provide trajectory-specific futures but require separate computation for every candidate. Shared predictions, such as the future occupancy used by ST-P3, amortize forecasting across candidates~\citep{hu2022stp3} but do not directly provide path-specific evidence. Extracting discriminative, trajectory-specific evidence from a shared forecast without repeated candidate-wise rollouts therefore remains a key challenge.

\subsection{Trajectory Refinement and Selection}

Candidate-based planning requires both reliable discrimination and local geometric improvement. VADv2 models a distribution over a discrete planning vocabulary, while Hydra-MDP distills human and rule-based teachers into metric-specific trajectory heads~\citep{jiang2026vadv2,li2024hydramdp}. DTPP learns context-dependent costs, Gen-Drive learns a candidate evaluator, and DriveSuprim improves hard-negative discrimination through coarse-to-fine filtering and self-distillation~\citep{huang2024dtpp,huang2025gendrive,yao2025drivesuprim}. For trajectory refinement, ThinkTwice and ResWorld use predicted future information to update an initial plan~\citep{jia2023thinktwice,zhang2026resworld}. These lines of work separately strengthen candidate evaluation and trajectory refinement, but their coordination within a common hypothesis pool remains less explored. Retaining unrefined alternatives and directly comparing retained and refined trajectories under the same future-conditioned evaluator thus remains an open design problem.

\section{Method}

Figure~\ref{fig:method_overview} outlines ManeuverFlow.
M-GMP allocates candidates across scene-relevant components, then M$^3$F transports their samples into bases in one step.
Metric Prism pre-scores each base using path-specific current/future BEV evidence.
ECFR replaces selected bases; refinements and retained bases re-query the field for joint scoring.

\subsection{Problem Formulation}

Let $\chi\in\{\mathrm{TF},\mathrm{LTF}\}$ denote the perception configuration fixed for each model. At planning step $n$, the planner receives
\begin{equation}
\mathcal O_n^\chi=
\begin{cases}
(I_n,L_n,s_n,c_n^{\mathrm{nav}}), & \chi=\mathrm{TF},\\
(I_n,s_n,c_n^{\mathrm{nav}}), & \chi=\mathrm{LTF},
\end{cases},
\end{equation}
where $I_n$ denotes the three camera views, $L_n$ is a LiDAR BEV raster, and $s_n$ and $c_n^{\mathrm{nav}}$ are the ego state and navigation command. Given $\mathcal O_n^\chi$, the planner predicts a future ego trajectory $\tau_n^\star$ with $H_\tau$ waypoints.

Each $H_\tau$-waypoint trajectory hypothesis $\tau_{n,i}=\{(x_{n,i,h},y_{n,i,h},\psi_{n,i,h})\}_{h=1}^{H_\tau}$ is represented in the current ego frame.
M$^3$F produces $N$ bases $\mathcal T_n^{\mathrm{base}}$. Metric Prism pre-scores them; the Refiner replaces $R$ selected bases with refinements and retains the other $N-R$ in the fixed-size pool $\mathcal T_n^{\mathrm{final}}$.
Metric Prism re-scores $\mathcal T_n^{\mathrm{final}}$ with trajectory-conditioned evidence and returns $\tau_n^\star$.

\subsection{Planning Context and M-GMP}

Using $L_n$ in TF and a learned $\mathbf Z_{\mathrm{lat}}$ in LTF, the encoder $\mathcal E_\chi$ returns the BEV map $B_n$, agent/maneuver queries $(Q_n^A,q_n^{\mathrm{man}})$, and state--navigation context $c_n^{\mathrm{SN}}$:
\begin{equation}
(B_n,Q_n^A,q_n^{\mathrm{man}},c_n^{\mathrm{SN}})
=\mathcal E_\chi(\mathcal O_n^\chi).
\end{equation}

The scene-conditioned maneuver weights are
\begin{equation}
\pi_n^{\mathrm{man}}
=\operatorname{Softmax}\!\left(
\mathcal D_{\mathrm{man}}(q_n^{\mathrm{man}},c_n^{\mathrm{SN}})
\right),
\end{equation}
\nopagebreak[4]
while all candidates share \mbox{$C_n=(B_n,Q_n^A,c_n^{\mathrm{SN}})$}.

Map and agent branches provide auxiliary supervision for BEV semantics and traffic-participant detection.
M-GMP partitions expert trajectories into $M$ motion families using displacement, mean heading, and speed trend, then fits a fixed factorized Gaussian to each family in normalized diff-space (Appendix). We use ``maneuver'' and ``intention'' to denote these trajectory-space motion components.
At inference, the scene-dependent weights $\pi_n^{\mathrm{man}}$ allocate candidates among components; selected component samples initialize the M$^3$F source states $e_{n,i}$.
The auxiliary objective is
\begin{equation}
\mathcal L_{\mathrm{aux}}
=\mathcal L_{\mathrm{map}}+\mathcal L_{\mathrm{agent}}+\mathcal L_{\mathrm{man}}.
\end{equation}
Its maneuver-classification term is
\begin{equation}
\mathcal L_{\mathrm{man}}
=\operatorname{CE}\!\left(y_n^{\mathrm{man}},\pi_n^{\mathrm{man}}\right),
\end{equation}
where $y_n^{\mathrm{man}}=\operatorname{Softmax}(-d_n)$ and $d_n$ collects expert--component distances in normalized diff-space (Appendix).

\begin{figure}[!t]
    \centering
    \includegraphics[width=\columnwidth]{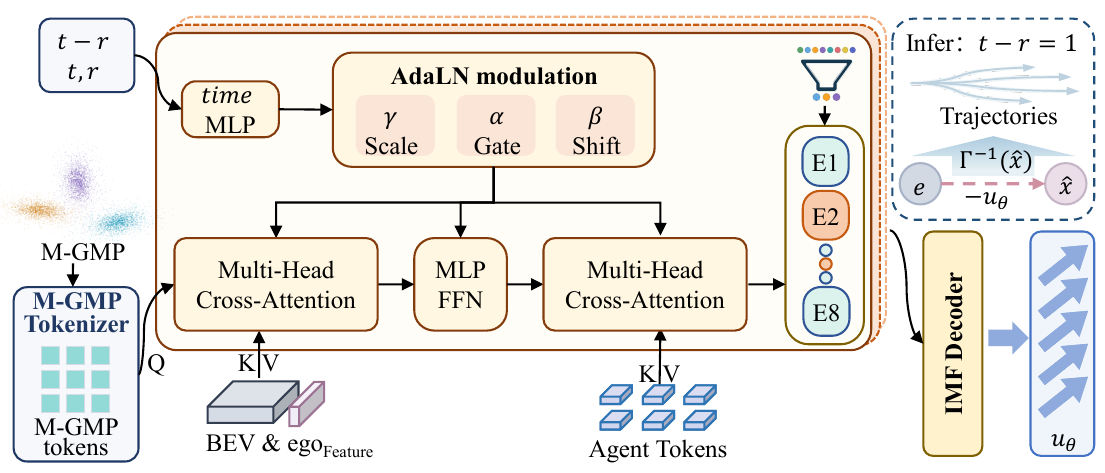}
    \caption{Architecture of M$^3$F. Interpolated trajectory-state tokens first query BEV and ego features, then agent memories. Time-conditioned AdaLN modulates each block, while maneuver-routed experts parameterize the interval-averaged field for one-step trajectory transport.}
    \label{fig:m3f_architecture}
\end{figure}

\subsection{M$^3$F}

M$^3$F adopts the boundary-field and JVP construction of Improved Mean Flows (iMF)~\citep{geng2025improved}; our driving-specific contributions are motion-structured M-GMP sources, scene-conditioned component allocation, and maneuver-routed experts.
The Appendix defines the trajectory encoding, maneuver assignment, and component-wise source sampling.
With expert code $x_n^{\mathrm{gt}}$ and source state $e_{n,i}$, define $z_{n,i}^{t}=(1-t)x_n^{\mathrm{gt}}+t e_{n,i}$ and $v_{n,i}^{c}=e_{n,i}-x_n^{\mathrm{gt}}$ for $0\leq r\leq t\leq1$.
Let $m_{n,i}$ index the M-GMP maneuver component generating $e_{n,i}$ and set $w_{n,i}^{\mathrm{man}}=\operatorname{OneHot}(m_{n,i})$; routing for the M$^3$F generation objective uses the expert's distance-matched motion component.

As shown in Fig.~\ref{fig:m3f_architecture}, the M-GMP tokenizer maps $z_{n,i}^{t}$ to the query $Q_{n,i}^{0}$.
The interval encoding $\varphi_{\mathrm F}(t,t-r)$ modulates the blocks, whose state queries sequentially attend to the BEV--ego and agent memories.
The maneuver code selects the corresponding feed-forward expert.
The decoded field is
\begin{equation}
\begin{aligned}
u_{\theta,n,i}
&\equiv u_\theta(z_{n,i}^{t},r,t\mid C_n,w_{n,i}^{\mathrm{man}})\\
&=\mathcal D_u\!\left(
\mathcal F_\theta^{\mathrm{M^3F}}
\left(Q_{n,i}^{0};\varphi_{\mathrm F}(t,t-r),C_n,w_{n,i}^{\mathrm{man}}\right)
\right).
\end{aligned}
\end{equation}

Fixing $(C_n,w_{n,i}^{\mathrm{man}})$, the boundary velocity at $(z_{n,i}^{t},t,t)$ supplies the JVP tangent, while the $(z,r,t)$-JVP is evaluated at the state and time coordinates $(z_{n,i}^{t},r,t)$:
\begin{gather}
v_{\theta,n,i}^{b}
=u_\theta(z_{n,i}^{t},t,t\mid C_n,w_{n,i}^{\mathrm{man}}),\\
J_{\theta,n,i}
=\operatorname{JVP}\!\left[
u_\theta;(z_{n,i}^{t},r,t);(v_{\theta,n,i}^{b},0,1)
\right],\\
\widetilde v_{\theta,n,i}
=u_{\theta,n,i}+(t-r)\operatorname{sg}[J_{\theta,n,i}],
\end{gather}
where $\operatorname{sg}[\cdot]$ denotes stop-gradient.
The loss is
\begin{equation}
\mathcal L_{\mathrm{M^3F}}
=\mathbb E\!\left[
\left\|\widetilde v_{\theta,n,i}-v_{n,i}^c\right\|_1
\right].
\end{equation}

\subsection{Future BEV Field and Metric Prism}

The world model predicts once per planning step a candidate-independent $H_B$-step future BEV feature field, providing prospective evidence without candidate-wise rollouts.
Write each state as $\widehat{\mathbf B}_{n,j}=(\widehat B_{n,j}^{\mathrm{abs}},\widehat B_{n,j}^{\mathrm{res}})$, where the residual is computed from consecutive absolute features with $\widehat B_{n,0}^{\mathrm{abs}}=B_n$:
\begin{equation}
\widehat{\mathcal F}_n
=\mathcal W_{\vartheta_{\mathrm W}}(B_n,c_n^{\mathrm{SN}})
=\{\widehat{\mathbf B}_{n,j}\}_{j=1}^{H_B}.
\end{equation}

For each candidate $\tau_{n,i}$, the learned Trajectory-Guided BEV Feature Probe projects its waypoints into the current BEV map and time-aligned predicted states, samples along the path, and aggregates trajectory-specific evidence. Thus, specificity comes from trajectory-conditioned retrieval over the shared forecast, not candidate-wise future prediction.
Metric Prism encodes the candidates as score tokens $\mathbf Q_n^S=[\mathcal T_S(\tau_{n,i})]_{i=1}^{N}$, compares them through set-level self-attention, and lets each token cross-attend to its associated evidence memory, yielding $q_{n,i}^S$.
Factor-specific heads produce $s_{n,i}^f=\mathcal D_f(q_{n,i}^S)$ and collect them into $\mathbf s_{n,i}$.
The evidence extraction and score composition are compactly written as
\begin{equation}
E_{n,i}=\mathcal P_\zeta\!\left(\tau_{n,i};B_n,\widehat{\mathcal F}_n\right),
\quad
S_{n,i}=\mathcal C_{\mathrm{PDM}}(\mathbf s_{n,i}),
\end{equation}
where $\mathbf s_{n,i}=\{s_{n,i}^f\}_{f\in\mathcal J_{\mathrm{PDM}}}$ contains the five predicted PDM factors in $\mathcal J_{\mathrm{PDM}}=\{\mathrm{NC},\mathrm{DAC},\mathrm{EP},\mathrm{TTC},\mathrm{Comf}\}$. The fixed log-PDM composition $\mathcal C_{\mathrm{PDM}}$ ranks candidates on both NAVSIM v1 and v2; v2-only factors remain evaluation metrics and do not enter selection.
The score $S_{n,i}$ ranks candidates, while $E_{n,i}$ conditions the Refiner.

During training, the shared perception backbone encodes current observations on the online path and is reused to encode future observations in a stop-gradient target pass.
After alignment to the current ego frame, the detached future features define $(\bar B_{n,j}^{\mathrm{abs}},\bar B_{n,j}^{\mathrm{res}})$, with $\Omega_{n,j}$ marking the valid spatial overlap.
Under LTF, this target pass processes three future camera views with the learned latent tensor in place of LiDAR.
The $\Omega_{n,j}$-masked future-BEV loss combines Smooth-$L_1$ and cosine alignment for absolute features with Smooth-$L_1$ supervision for temporal residuals. For each \texttt{navtrain} scene, Metric Prism learns from $N=64$ candidates and their NAVSIM v1 evaluator factor labels using weighted binary cross-entropy.
Summing over $j=1,\ldots,H_B$, $i=1,\ldots,N$, and
$f\in\mathcal J_{\mathrm{PDM}}$, the overall verification objective is
\begin{equation}
\begin{aligned}
\mathcal L_{\mathrm{verify}}
&=\mathcal L_{\mathrm{FBEV}}+\mathcal L_{\mathrm{score}}\\
&=\sum_j\!\left(\ell_{n,j}^{\mathrm{abs}}+\beta\ell_{n,j}^{\mathrm{res}}\right)
+\sum_{i,f}\ell_f\!\left(s_{n,i}^f,y_{n,i}^f\right),
\end{aligned}
\end{equation}
where $\beta$ weights residual supervision, $y_{n,i}^f$ is the evaluator label, and $\ell_f$ includes factor weighting, label smoothing, and logit regularization, detailed in the supplement.

\subsection{ECFR}

Given a base candidate $\tau_{n,i}$ and its evidence memory $E_{n,i}$, the Refiner predicts a residual in the path-aligned Frenet frame and updates the trajectory as
\begin{equation}
\Delta_{n,i}^{\mathrm F}
 = R_\eta\!\left(\tau_{n,i},E_{n,i}\right),
\qquad
\widetilde{\tau}_{n,i}
 = \mathcal U_{\mathrm F}\!\left(\tau_{n,i},\Delta_{n,i}^{\mathrm F}\right).
\end{equation}
During training, the same $N=64$ set $\mathcal T_n^{\mathrm{base}}$ supplies both score labels and refinement targets. For each base, the highest evaluator-scoring candidate with the same maneuver index defines its reference trajectory:
\begin{gather}
\tau_{n,i}^\dagger
=\operatorname*{arg\,max}_{\substack{
\tau_{n,j}\in\mathcal T_n^{\mathrm{base}}\\
m_{n,j}=m_{n,i}}}
P_{\mathrm{eval}}(\tau_{n,j}),\\
\mathcal L_{\mathrm{ref}}
=\frac{\alpha_{\mathrm{ref}}\sum_i\omega_{n,i}\,
\ell_{\mathrm F}\!\left(\Delta_{n,i}^{\mathrm F},\bar\Delta_{n,i}^{\mathrm F}\right)}
{\max(1,\sum_i\omega_{n,i})},
\end{gather}
where $P_{\mathrm{eval}}$ combines evaluator factor labels, $\bar\Delta_{n,i}^{\mathrm F}=\mathcal R_{\mathrm F}(\tau_{n,i},\tau_{n,i}^\dagger)$ is the Frenet target, and $\ell_{\mathrm F}$ is mean Smooth-$L_1$. We use $\alpha_{\mathrm{ref}}=5$ and binary validity weights $\omega_{n,i}$; invalid pairs use zero targets and weights.

\subsection{Training Objectives and Inference}

Training proceeds in two stages.
We first optimize the auxiliary maneuver objective $\mathcal L_{\mathrm{aux}}$ and the M$^3$F generation objective $\mathcal L_{\mathrm{M^3F}}$.
Stage~2 jointly optimizes all inference-time modules. At each training step, the current model generates 64 candidates per scene for evaluator labeling and ECFR reference selection. Candidate construction and supervision targets are detached:
\begin{equation}
\begin{aligned}
\mathcal L_{\mathrm{joint}}
&=\lambda_{\mathrm{aux}}\mathcal L_{\mathrm{aux}}
+\lambda_{\mathrm{M^3F}}\mathcal L_{\mathrm{M^3F}}\\
&\quad+\lambda_{\mathrm{verify}}\mathcal L_{\mathrm{verify}}
+\lambda_{\mathrm{ref}}\mathcal L_{\mathrm{ref}},
\end{aligned}
\end{equation}
where the $\lambda$ coefficients balance the four objectives, while $\mathcal L_{\mathrm{verify}}$ combines future-BEV prediction and factor-score supervision as defined in Sec.~3.4.

At inference, we retain the Top-3 maneuver components of $\pi_n^{\mathrm{man}}$, renormalize their mixture weights, and distribute $N$ candidates by largest-remainder allocation.
For each selected component $m$, we sample $\widetilde e_{n,i}$ from its displacement--heading Gaussian, set $e_{n,i}=g(\widetilde e_{n,i})$, and assign $m_{n,i}=m$ and $w_{n,i}^{\mathrm{man}}=\operatorname{OneHot}(m)$.
Each prior is transported into a trajectory through one full-interval M$^3$F evaluation:
\begin{equation}
\tau_{n,i}
=\Gamma^{-1}\!\left(
e_{n,i}-u_\theta(e_{n,i},0,1\mid C_n,w_{n,i}^{\mathrm{man}})
\right),
\end{equation}
where $\Gamma^{-1}$ applies $\mathcal A^{-1}$, cumulatively sums displacements from the current ego origin, and decodes heading with $\operatorname{atan2}$.

Metric Prism first pre-scores the $N$ base candidates.
Guided by the preliminary ranking, selected bases are replaced with Refiner outputs that inherit their maneuver indices as provenance, while the remaining candidates are retained.
For the resulting fixed-size pool $\mathcal T_n^{\mathrm{final}}$, trajectory evidence is re-queried from the shared future field before Metric Prism jointly re-scores all candidates under the deployment log-PDM composition.
The highest-scoring trajectory is returned as the final plan $\tau_n^\star$.

\section{Experiments}

\begin{table*}[t]
\centering
\small
\setlength{\tabcolsep}{3.5pt}
\begin{tabular}{@{}l l c c c c c c c@{}}
\toprule
Method & Venue & Backbone & NC$\uparrow$ & DAC$\uparrow$ & TTC$\uparrow$ & Comf.$\uparrow$ & EP$\uparrow$ & PDMS$\uparrow$ \\
\midrule
UniAD~\citep{hu2023planning} & CVPR 2023 & ResNet-34 & 97.8 & 91.9 & 92.9 & 100.0 & 78.8 & 83.4 \\
TransFuser$^{\mathrm{TF}}$~\citep{chitta2022transfuser} & TPAMI 2023 & ResNet-34 & 97.7 & 92.8 & 92.8 & 100.0 & 79.2 & 84.0 \\
World4Drive~\citep{zheng2025world4drive} & ICCV 2025 & ResNet-34 & 97.4 & 94.3 & 92.8 & 100.0 & 79.9 & 85.1 \\
GoalFlow$^{\mathrm{TF}}$~\citep{xing2025goalflow} & CVPR 2025 & ResNet-34 & 98.3 & 93.8 & 94.3 & 100.0 & 79.8 & 85.7 \\
DiffusionDrive$^{\mathrm{TF}}$~\citep{liao2025diffusiondrive} & CVPR 2025 & ResNet-34 & 98.2 & 96.2 & 94.7 & 100.0 & 82.2 & 88.1 \\
WoTE$^{\mathrm{TF}}$~\citep{li2025wote} & ICCV 2025 & ResNet-34 & 98.5 & 96.8 & 94.9 & 99.9 & 81.9 & 88.3 \\
MeanFuser$^{\mathrm{LTF}}$~\citep{wang2026meanfuser} & CVPR 2026 & ResNet-34 & 98.6 & 97.0 & 95.0 & 100.0 & 82.8 & 89.0 \\
VADv2~\citep{jiang2026vadv2} & ICLR 2026 & ResNet-34 & 98.3 & 97.4 & 95.7 & 100.0 & 82.3 & 89.3 \\
DriveSuprim~\citep{yao2025drivesuprim} & AAAI 2026 & ResNet-34 & 97.8 & 97.3 & 93.6 & 100.0 & 86.7 & 89.9 \\
\midrule
\rowcolor{black!6} Ours$^{\mathrm{LTF}}$ & -- & ResNet-34 & 98.49 & 97.40 & 95.13 & 99.99 & 84.87 & \textbf{89.98} \\
\rowcolor{black!6} Ours$^{\mathrm{TF}}$ & -- & ResNet-34 & 98.42 & 98.01 & 95.19 & 99.99 & 85.36 & \textbf{90.58} \\
\bottomrule
\end{tabular}
\caption{NAVSIM v1 navtest mean results under a single official benchmark version; baselines use official results. Standard deviations are reported in the supplementary material.}
\label{tab:navsim-v1-pdms}
\end{table*}

\begin{table*}[t]
\centering
\small
\setlength{\tabcolsep}{2.5pt}
\begin{tabular}{@{}l c c c c c c c c c c@{}}
\toprule
Method & NC$\uparrow$ & DAC$\uparrow$ & DDC$\uparrow$ & TLC$\uparrow$ & EP$\uparrow$ & TTC$\uparrow$ & LK$\uparrow$ & HC$\uparrow$ & EC$\uparrow$ & EPDMS$\uparrow$ \\
\midrule
Ego Status MLP~\citep{dauner2024navsim} & 93.1 & 77.9 & 92.7 & 99.6 & 86.0 & 91.5 & 89.4 & 98.3 & 85.4 & 64.0 \\
TransFuser$^{\mathrm{TF}}$~\citep{chitta2022transfuser} & 96.9 & 89.9 & 97.8 & 99.7 & 87.1 & 95.4 & 92.7 & 98.3 & 87.2 & 76.7 \\
DriveSuprim~\citep{yao2025drivesuprim} & 97.5 & 96.5 & 99.4 & 99.6 & 88.4 & 96.6 & 95.5 & 98.3 & 77.0 & 83.1 \\
DiffusionDrive$^{\mathrm{TF}}$~\citep{liao2025diffusiondrive} & 98.2 & 96.3 & 99.4 & 99.8 & 87.4 & 97.4 & 97.0 & 98.3 & 87.7 & 88.3 \\
MeanFuser$^{\mathrm{LTF}}$~\citep{wang2026meanfuser} & 98.3 & 97.2 & 99.6 & 99.8 & 87.6 & 97.4 & 97.3 & 98.3 & 88.2 & 89.5 \\
\midrule
\rowcolor{black!6} Ours$^{\mathrm{LTF}}$ & 98.61 & 98.13 & 99.25 & 99.80 & 87.27 & 97.96 & 96.13 & 98.26 & 85.17 & \textbf{89.70} \\
\rowcolor{black!6} Ours$^{\mathrm{TF}}$ & 98.38 & 97.94 & 99.41 & 99.76 & 88.75 & 97.75 & 96.86 & 98.32 & 86.54 & \textbf{90.14} \\
\bottomrule
\end{tabular}
\caption{NAVSIM v2 navtest mean results under a single official benchmark version; baselines use official results. Standard deviations are reported in the supplementary material.}
\label{tab:navsim-v2-epdms}
\end{table*}

\begin{table}[!t]
\centering
{
\small
\setlength{\tabcolsep}{2.0pt}
\begin{tabular}{@{}l c c c c@{}}
\toprule
Method & Traffic & SR$\uparrow$ & EP$\uparrow$ & PDMS$^\ast\uparrow$ \\
\midrule
VADv2~\citep{jiang2026vadv2} & Log & 77.08 & 48.00 & 63.81 \\
VADv2~\citep{jiang2026vadv2} & IDM & 73.66 & 46.71 & 60.98 \\
\addlinespace[1pt]
DiffusionDrive~\citep{liao2025diffusiondrive} & Log & 86.24 & 45.09 & 69.83 \\
DiffusionDrive~\citep{liao2025diffusiondrive} & IDM & 85.85 & 43.88 & 68.22 \\
\addlinespace[1pt]
MeanFuser~\citep{wang2026meanfuser} & Log & 87.25 & 55.90 & 73.81 \\
MeanFuser~\citep{wang2026meanfuser} & IDM & 87.15 & 55.60 & 73.28 \\
\midrule
\rowcolor{black!6} \textbf{ManeuverFlow} & Log & 87.48 & 60.68 & \textbf{75.19} \\
\rowcolor{black!6} \textbf{ManeuverFlow} & IDM & 87.92 & 60.89 & \textbf{74.58} \\
\bottomrule
\end{tabular}
}
\caption{Closed-loop results on all WorldEngine scenarios.}
\label{tab:worldengine}
\end{table}

\subsection{Experimental Setup}

\paragraph{Datasets and evaluation protocols.}
NAVSIM is a data-driven planning benchmark built from OpenScene/nuPlan driving logs~\citep{karnchanachari2024nuplan} that evaluates predicted ego trajectories against logged future scenes under a non-reactive simulator protocol~\citep{dauner2024navsim,cao2025pseudosimulation}. We follow the official split: \texttt{navtrain} for training and \texttt{navtest} for evaluation. We estimate M-GMP motion families and distribution statistics using only expert trajectories from \texttt{navtrain}. NAVSIM v1 reports the Predictive Driver Model Score (PDMS) with NC, DAC, TTC, EP, and Comfort, whereas v2 adopts the Extended Predictive Driver Model Score (EPDMS) and additionally reports DDC, TLC, LK, HC, and EC. We use the official evaluators; for v2, we reuse the v1 checkpoints, five-factor heads, and log-PDM rule---without v2-specific training or tuning.

We complement NAVSIM's non-reactive evaluation with the latest WorldEngine benchmark on all 288 publicly released high-risk scenarios reconstructed from NAVSIM \texttt{navtest} logs~\citep{li2026worldengine}. The logged replay (Log) setting replays background traffic, whereas the Intelligent Driver Model (IDM) setting updates surrounding agents reactively~\citep{treiber2000idm}; in both settings, the ego vehicle replans from updated rendered observations. We report episode success rate (SR), closed-loop ego progress (EP), and PDMS$^\ast$: SR is the fraction of episodes without collisions or drivable-area violations, EP measures progress during closed-loop execution, and PDMS$^\ast$ summarizes overall closed-loop driving quality, together assessing system-level replanning under replayed and reactive traffic.

\begin{table*}[!t]
\centering
\small
\setlength{\tabcolsep}{3.5pt}
\begin{tabular}{@{}l l c c c c c c c@{}}
\toprule
Model & Components & Prior & Flow & Evidence & Refiner
& NC$\uparrow$ & EP$\uparrow$ & PDMS$\uparrow$ \\
\midrule
Base
& Direct planner
& -- & -- & -- & -- & 97.73 & 79.24 & 84.01 \\

$\mathcal{M}_0$
& Candidate-based MeanFlow
& Single & Dense MF & Cur. & --
& 97.21 & 81.05 & 86.14 \\

$\mathcal{M}_1$
& + M-GMP and Dense-XL
& M-GMP-8 & Dense-XL & Cur. & --
& 97.02 & 81.63 & 87.91 \\

$\mathcal{M}_2$
& + maneuver-routed flow
& M-GMP-8 & M$^3$F & Cur. & --
& 97.64 & 82.12 & 88.35 \\

$\mathcal{M}_3$
& + future evidence
& M-GMP-8 & M$^3$F & Cur.+Fut. & --
& 98.52 & 84.41 & 89.93 \\

\midrule
$\mathcal{M}_4^{\mathrm{Cur}}$
& w/o future branch
& M-GMP-8 & M$^3$F & Cur. & ECFR
& 98.27 & 84.50 & 90.03 \\

$\mathcal{M}_4^{\mathrm{KM}}$
& KMeans prior control
& KMeans-8 & M$^3$F & Cur.+Fut. & ECFR
& 98.42 & 84.53 & 90.04 \\

\rowcolor{black!6} $\mathcal{M}_4$
& Complete model
& M-GMP-8 & M$^3$F & Cur.+Fut. & ECFR
& 98.42 & 85.36 & \textbf{90.58} \\
\bottomrule
\end{tabular}
\caption{Cumulative ablations with matched full-pipeline controls on NAVSIM v1 navtest (standard TF). Dense-XL is parameter-count matched to M$^3$F; $\mathcal{M}_4^{\mathrm{Cur}}$ disables the predicted-future branch, while $\mathcal{M}_4^{\mathrm{KM}}$ replaces M-GMP-8 with KMeans-8. Cur./Fut. = current/predicted-future BEV evidence.}
\label{tab:main_ablation}
\end{table*}

\paragraph{Implementation details.}
TF and LTF use a ResNet-34 image backbone and predict $H_\tau=8$ waypoints over 4\,s. We conduct three independent training runs on four NVIDIA RTX 4090 GPUs with global batch size 32 using AdamW (weight decay 0.1). Stage one trains the perception network and M$^3$F at $4\times10^{-4}$; stage two jointly optimizes all inference-time modules at $2\times10^{-4}$, with a $0.2\times$ learning-rate multiplier for the perception network and M$^3$F. The future field predicts $H_B=4$ BEV states at 1-s intervals; only the stop-gradient target branch uses future observations during training, while inference uses the current observation. The defaults $k=3$, $N=64$, and $R=N/4$ were fixed before \texttt{navtest}; all \texttt{navtest} sweeps are post-hoc sensitivity analyses using frozen checkpoints. Each checkpoint is evaluated five times, and the main tables report the mean across the three independent run averages; the corresponding sample standard deviations are provided in the supplementary material.

\subsection{Main Results}

\paragraph{Results on NAVSIM.}
ManeuverFlow-TF achieves the highest listed PDMS of 90.58 in Table~\ref{tab:navsim-v1-pdms}, while the camera-only LTF variant obtains 89.98. Their EP values are 85.36 and 84.87, respectively.

Without v2 adaptation, ManeuverFlow-TF achieves the highest listed EPDMS and EP at 90.14 and 88.75. ManeuverFlow-LTF obtains 89.70 EPDMS and the highest listed NC, DAC, and TTC at 98.61, 98.13, and 97.96 (Table~\ref{tab:navsim-v2-epdms}). Thus, the v1-trained checkpoints remain competitive under v2's broader metric suite without optimization for its additional v2 evaluation factors.

\paragraph{Closed-loop results in WorldEngine.}
WorldEngine provides 3DGS-rendered camera observations but no LiDAR input; Table~\ref{tab:worldengine} therefore evaluates only ManeuverFlow-LTF. Under Log and IDM, it obtains the highest listed PDMS$^\ast$ values, 75.19 and 74.58, respectively. Its EP values of 60.68 and 60.89 are also the highest listed, while SR reaches 87.48 and 87.92. These results complement NAVSIM with closed-loop evidence for ego progress and composite performance under replayed and reactive traffic.

\begin{figure*}[!t]
    \centering
\includegraphics[width=\textwidth]{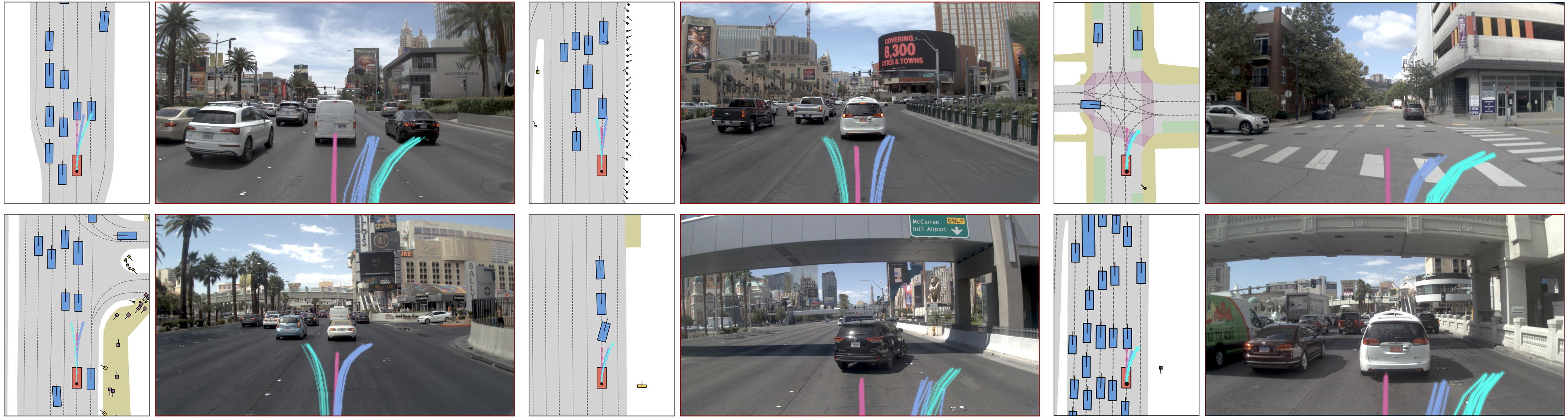}
    \caption{Six urban scenes with the top-three maneuver components shown in paired BEV and front-view projections.}
    \label{fig:qualitative_trajectories}
\end{figure*}

\subsection{Ablation Studies}
\label{sec:ablation}

Table~\ref{tab:main_ablation} decomposes the Hypothesize--Verify--Improve pipeline under standard TF; $\mathcal{M}_4^{\mathrm{Cur}}$ removes the predicted-future branch, while $\mathcal{M}_4^{\mathrm{KM}}$ changes only the prior.

\emph{Hypothesize.} Candidate-based MeanFlow improves PDMS by 2.13 points. $\mathcal{M}_0\!\to\!\mathcal{M}_1$ jointly adds M-GMP-8 and the parameter-matched Dense-XL generator for another 1.77 points; because both prior support and capacity change, this is stage-level evidence. With M-GMP-8 fixed, M$^3$F adds 0.44 PDMS over the parameter-count-matched Dense-XL control, so total parameter count alone does not explain the gain. In the complete pipeline, M-GMP-8 exceeds KMeans-8 by 0.54 PDMS and 0.83 EP, supporting maneuver-aligned prior organization. Appendix Figs.~S1--S2 provide the corresponding prior visualizations.

\emph{Verify.} Before refinement, predicted-future evidence adds 1.58 PDMS and 2.29 EP at a fixed candidate budget. The gain persists after ECFR: relative to $\mathcal{M}_4^{\mathrm{Cur}}$, the complete model adds 0.55 PDMS and 0.86 EP under matched protocols for candidate generation and evaluator supervision.

\emph{Improve.} ECFR adds 0.65 PDMS and 0.95 EP to the future-aware pipeline, with NC decreasing by 0.10 points, and also improves the current-only variant. Future-aware verification and ECFR yield complementary but sub-additive gains.

\paragraph{M-GMP recall and proposal-budget sensitivity.}
Table~\ref{tab:maneuver_budget} reports M-GMP recall@$k$, which measures whether the target component appears among the top-$k$ routed components. Recall reaches 98.30\% at $k=3$ and changes marginally thereafter. With $k=3$, PDMS saturates beyond $N=64$ as the activation bound rises and Proposal FPS falls, supporting $N=64$ as a balanced operating point.

Taken together, these ablations support ManeuverFlow's intended lifecycle: maneuver routing organizes a compact candidate set, shared-future evidence improves planning performance, and evidence-conditioned ECFR improves final-plan quality through maneuver-matched refinement.

\begin{table}[!t]
\centering
\small
\setlength{\tabcolsep}{2.8pt}
\begin{tabular*}{\columnwidth}{@{\extracolsep{\fill}}c c r r r r r@{}}
\toprule
\multicolumn{2}{c}{Component routing}
& \multicolumn{5}{c}{PDMS$\uparrow$ at proposal budget $N$} \\
\cmidrule(lr){1-2}
\cmidrule(lr){3-7}
$k$ & \shortstack{M-GMP target\\recall@$k$ (\%)} & 8 & 16 & 32 & 64 & 128 \\
\midrule
1 & 87.20 & 89.85 & 90.04 & 90.26 & 90.33 & 90.42 \\
2 & 96.46 & 90.04 & 90.15 & 90.43 & 90.54 & 90.59 \\
3 & 98.30 & 90.06 & 90.28 & 90.41 & 90.58$^\dagger$ & 90.56 \\
4 & 99.21 & 90.10 & 90.27 & 90.42 & 90.57 & 90.67 \\
\midrule
\multicolumn{2}{l}{M$^3$F activ. bound (MB) $\downarrow$}
& 1.11 & 2.07 & 3.99 & 7.82 & 15.50 \\
\multicolumn{2}{l}{Proposal FPS $\uparrow$}
& 66.46 & 63.69 & 58.62 & 53.86 & 47.12 \\
\bottomrule
\end{tabular*}
\caption{M-GMP routing and proposal-budget sensitivity on NAVSIM v1 navtest; FPS: perception--M$^3$F; $^\dagger$: default.}
\label{tab:maneuver_budget}
\end{table}

\subsection{Qualitative Analysis}
\label{sec:qualitative}

\paragraph{Maneuver-structured hypothesis generation.}
Figure~\ref{fig:qualitative_trajectories} shows M$^3$F samples sharing a coherent motion tendency within each routed M-GMP component while spanning distinct, road-aligned lane and intersection alternatives across components. In these examples, the hypotheses remain visually consistent with drivable space and surrounding traffic, illustrating component-conditioned trajectory alternatives for downstream evaluation. Across the paired BEV and camera views, their separation remains legible around lane choices and intersections, making the routed maneuver structure directly inspectable. See the appendix for additional cross-method comparisons and visualizations of the maneuver-component priors before trajectory transport.

\section{Conclusion}
We present \textbf{ManeuverFlow}, an explicit \emph{Hypothesize--Verify--Improve} planner that maintains maneuver-indexed hypotheses throughout generation, evaluation, and refinement. \mbox{M-GMP} allocates a fixed proposal budget across scene-relevant motion components, and M$^3$F transports their samples into scene-adaptive, maneuver-routed hypotheses in one step. A shared predicted future BEV field provides path-specific evidence for Metric Prism scoring and ECFR's local Frenet refinement. Refined and retained hypotheses re-query the shared field and are jointly re-scored for final selection. Complementary non-reactive and closed-loop evaluations on NAVSIM and WorldEngine, respectively, demonstrate the \mbox{effectiveness} of the integrated planner.

\bibliography{aaai2027}

\clearpage
\onecolumn
\setcounter{section}{0}
\setcounter{subsection}{0}
\setcounter{equation}{0}
\setcounter{figure}{0}
\setcounter{table}{0}
\renewcommand{\theequation}{S\arabic{equation}}
\renewcommand{\thefigure}{S\arabic{figure}}
\renewcommand{\thetable}{S\arabic{table}}
\setlength{\textheight}{666pt}
\makeatletter
\global\@colht\textheight
\global\@colroom\textheight
\global\vsize\textheight
\let\@addtocurcol\MFappendixaddtocurcol
\setlength{\@fptop}{0pt}
\setlength{\@fpbot}{0pt plus 1fil}
\setlength{\@dblfptop}{0pt}
\setlength{\@dblfpbot}{0pt plus 1fil}
\makeatother
\setlength{\textfloatsep}{10pt plus 2pt minus 2pt}
\setlength{\dbltextfloatsep}{10pt plus 2pt minus 2pt}
\setlength{\floatsep}{8pt plus 2pt minus 2pt}
\setlength{\dblfloatsep}{8pt plus 2pt minus 2pt}
\captionsetup{skip=6pt}

\raggedbottom
\clubpenalty=10000
\widowpenalty=10000
\displaywidowpenalty=10000
\twocolumn[
\begin{center}
{\Large\bfseries Appendix\par}
\end{center}
\vspace{0.5em}
]

\section{Hypothesize: Operational Maneuver Prior and Routed Transport}
\label{app:mgmp_construction}

\subsection{Diff-Space and Prior Construction}
For an expert trajectory anchored at planning step $n$,
$\tau_n=\{(\mathbf p_{n,h},\psi_{n,h})\}_{h=1}^{H_\tau}$ in the current ego frame,
let $H_\tau=8$, $\mathbf p_{n,h}=(x_{n,h},y_{n,h})^{\mathsf T}\in\mathbb R^2$, and
$\mathbf p_{n,0}=\mathbf 0$. We first express each waypoint as an incremental
displacement,
\begin{equation}
\Delta\mathbf p_{n,h}
 =\mathbf p_{n,h}-\mathbf p_{n,h-1}.
\label{eq:app_step_displacement}
\end{equation}
The fixed affine calibration $\mathcal A$ then maps this displacement to the
normalized space shared with the trajectory generator:
\begin{equation}
\Delta\widetilde{\mathbf p}_{n,h}
 =\mathcal A(\Delta\mathbf p_{n,h}).
\label{eq:app_normalized_step}
\end{equation}
Appending the raw heading to the normalized displacement gives the three-dimensional pre-embedding code
\begin{equation}
\widetilde\Gamma(\tau_n)_h
 =[
\Delta\widetilde{\mathbf p}_{n,h}^{\mathsf T},
\psi_{n,h}
 ]^{\mathsf T}.
\label{eq:app_preembedding_code}
\end{equation}
Finally, the cyclic heading embedding yields the four-dimensional M$^3$F code
\begin{equation}
\begin{aligned}
\Gamma(\tau_n)_h
 &=g\!\left(\widetilde\Gamma(\tau_n)_h\right)\\
 &=[
\Delta\widetilde{\mathbf p}_{n,h}^{\mathsf T},
\sin\psi_{n,h},\cos\psi_{n,h}
].
\end{aligned}
\label{eq:app_m3f_code}
\end{equation}
The calibration of $\mathcal A$ remains fixed throughout. M-GMP fits its
factorized Gaussian in the normalized pre-embedding space
$\widetilde\Gamma(\tau_n)\in\mathbb R^{H_\tau\times3}$; $g$ yields the
M$^3$F code $\Gamma(\tau_n)\in\mathbb R^{H_\tau\times4}$. Thus the main-paper
states are $x_n^{\mathrm{gt}}=\Gamma(\tau_n^{\mathrm{gt}})$ and
$e_{n,i}=g(\widetilde e_{n,i})$.

For prior construction, the Maneuver Gaussian Mixture Prior (M-GMP) has
$M=8$ components, is fitted
to expert trajectories from \texttt{navtrain} and assigns each trajectory to a
rule-defined motion family $m_n^{\mathrm{rule}}=\Psi(\boldsymbol\xi_n)$ using
$\boldsymbol\xi_n=(D_n,\bar\psi_n,\Delta\bar s_n)$. The first-to-last
displacement magnitude is
\begin{equation}
D_n
 =\lVert\mathbf p_{n,H_\tau}-\mathbf p_{n,1}\rVert_2.
\label{eq:app_endpoint_displacement}
\end{equation}
We summarize the signed turning tendency by the horizon-mean heading in degrees,
\begin{equation}
\bar\psi_n
 =\frac{180}{\pi H_\tau}\sum_{h=1}^{H_\tau}\psi_{n,h}.
\label{eq:app_mean_heading}
\end{equation}
For the speed-trend proxy, the length of interval $h$ is
\begin{equation}
\ell_{n,h}
 =\lVert\mathbf p_{n,h+1}-\mathbf p_{n,h}\rVert_2,
 \qquad h=0,\ldots,7.
\label{eq:app_interval_length}
\end{equation}
The difference between the late and early four-step mean interval lengths is defined as
\begin{equation}
\Delta\bar s_n
 =\tfrac{1}{4}\!\left(
\sum_{h=4}^{7}\ell_{n,h}
-\sum_{h=0}^{3}\ell_{n,h}
\right).
\label{eq:app_speed_trend}
\end{equation}
Here, $D_n$ is the first-to-last distance, $\bar\psi_n$ is mean heading in
degrees, and $\ell_{n,0}$ includes the ego-to-first-waypoint interval. The
symmetric $\Delta\bar s_n$ compares mean step lengths over equal early and late
windows, so it is a speed-trend proxy rather than physical acceleration. We
fix $\eta_s=0.25$ m per interval; Table~\ref{tab:app_mgmp_families} gives the
hierarchy, with $D_n\ge5$ m for every non-stop motion family in the prior.

The thresholds in Table~\ref{tab:app_mgmp_families} use degrees. The Gaussian
heading variables $\psi$, $\mu^\psi$, and $\overline\sigma^\psi$, together with
$\operatorname{wrap}(\cdot)$ below, use radians.

The component names form the operational trajectory-space vocabulary for
``maneuver'' and ``intention'' in the main paper. Their meaning is grounded in
measurable displacement, turning, and speed-trend tendencies rather than in a
separate lane-topology annotation protocol. Thus, ``intention-specific''
denotes conditioning on a motion-level planning representation:
``Accelerating''/``Decelerating'' use $\Delta\bar s_n$, while lane-change/turn
names use mean heading.

For each family and horizon step, we estimate the displacement mean
$\boldsymbol\mu^p_{m,h}$, its diagonal positive standard deviation
$\overline{\boldsymbol\sigma}^p_{m,h}$, and the circular heading location and
positive scale $(\mu^\psi_{m,h},\overline\sigma^\psi_{m,h})$ from its assigned
expert trajectories. A sample from component $m$ has normalized displacement
\begin{equation}
\Delta\widetilde{\mathbf p}_{i,m,h}
 =\boldsymbol\mu^p_{m,h}
+ \overline{\boldsymbol\sigma}^p_{m,h}
 \odot\boldsymbol\epsilon^p_{i,h}.
\label{eq:app_sample_displacement}
\end{equation}
The corresponding heading sample is
\begin{equation}
\widehat\psi_{i,m,h}
 =\mu^\psi_{m,h}
 + \overline{\sigma}^\psi_{m,h}\epsilon^\psi_{i,h}.
\label{eq:app_sample_heading}
\end{equation}
Together they form the sampled pre-embedding state
\begin{equation}
\widetilde{\mathbf e}_{i,m,h}
 =[
\Delta\widetilde{\mathbf p}_{i,m,h}^{\mathsf T},
 \widehat\psi_{i,m,h}]^{\mathsf T}.
\label{eq:app_sample_preembedding}
\end{equation}
Applying the same cyclic map $g$ gives the M$^3$F source code
\begin{equation}
\begin{aligned}
\mathbf e_{i,m,h}
 &=g(\widetilde{\mathbf e}_{i,m,h})\\
 &=[
\Delta\widetilde{\mathbf p}_{i,m,h}^{\mathsf T},
\sin\widehat\psi_{i,m,h},\cos\widehat\psi_{i,m,h}
 ].
\end{aligned}
\label{eq:app_sample_code}
\end{equation}
Here, $\boldsymbol\epsilon^p_{i,h}\sim\mathcal N(\mathbf0,\mathbf I_2)$ and
$\epsilon^\psi_{i,h}\sim\mathcal N(0,1)$; barred scales denote fixed positive
component standard deviations.
Rules fix component identity; their statistics are estimated from the assigned
experts and then held fixed.
For a scene-routed component $m=m_{n,i}$, stacking the waypoint states over
$h=1,\ldots,H_\tau$ gives the main-paper source tensors
$\widetilde e_{n,i}=[\widetilde{\mathbf e}_{i,m,h}]_{h=1}^{H_\tau}$ and
$e_{n,i}=[\mathbf e_{i,m,h}]_{h=1}^{H_\tau}$; $g$ acts pointwise.

\begin{figure*}[!t]
\centering
\begin{minipage}[t]{0.48\textwidth}
\centering
\includegraphics[width=\linewidth]{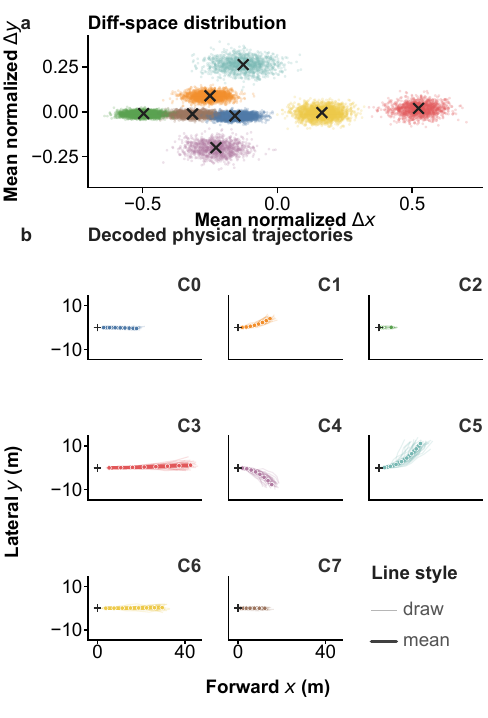}
\caption{KMeans-8 control prior before M$^3$F transport. \textbf{(a)} KMeans
components in mean normalized displacement space; crosses mark component
centers. \textbf{(b)} Decoded source trajectories for components C0--C7.}
\label{fig:app_kmeans_prior}
\end{minipage}
\hfill
\begin{minipage}[t]{0.48\textwidth}
\centering
\includegraphics[width=\linewidth]{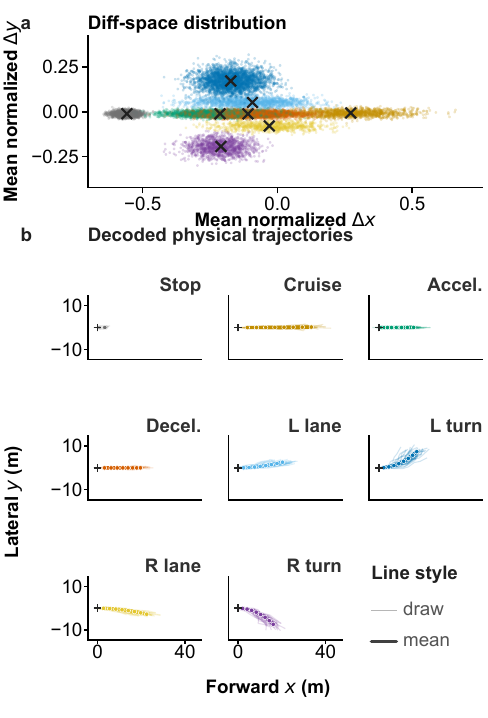}
\caption{Rule-aligned M-GMP-8 prior used by ManeuverFlow before M$^3$F
transport. \textbf{(a)} Expert trajectories organized by the trajectory-derived
maneuver families in Table~\ref{tab:app_mgmp_families}; crosses mark component means.
\textbf{(b)} Decoded source trajectories for the eight components.}
\label{fig:app_mgmp_prior}
\end{minipage}
\end{figure*}

Figures~\ref{fig:app_kmeans_prior} and~\ref{fig:app_mgmp_prior} visualize, on
matched axes, the two source priors compared in the complete-pipeline ablation:
the KMeans-8 control and the rule-aligned M-GMP-8 used by ManeuverFlow. Both
figures describe pre-transport source organization; the downstream M$^3$F and
the remaining pipeline are matched. Together, they are the two matched pre-transport prior
visualizations referenced in the main paper as ``Appendix Figs.~S1--S2.''

\paragraph{Evidence role of the overview schematic.}
Main-paper Fig.~1(a) is a mechanism-level schematic of the concentration risk
for query candidates without explicit mode structure, rather than a measured
collapse rate for a particular baseline. The empirical evaluation does not
infer an effect from the schematic. It evaluates the structured alternative
through the stage-level $\mathcal M_0\!\rightarrow\!\mathcal M_1$ change, the
parameter-count-matched $\mathcal M_1\!\rightarrow\!\mathcal M_2$ generator
control, the matched $\mathcal M_4^{\mathrm{KM}}\!\rightarrow\!\mathcal M_4$
prior control, and the target-component Recall@$k$/proposal-budget analysis in
the main paper. Figures~\ref{fig:app_kmeans_prior}--\ref{fig:app_mgmp_prior}
show the matched source-prior organization, while the main-paper qualitative
figure and Fig.~\ref{fig:app_m3f_expert_specialization} show routed
post-transport behavior. The schematic therefore motivates the failure mode;
the quantitative controls and diagnostics test the proposed response to it.

\begin{table}[!b]
\centering
{\small
\setlength{\tabcolsep}{2.0pt}
\begin{tabular}{@{}c
  >{\raggedright\arraybackslash}p{0.23\columnwidth}
  >{\raggedright\arraybackslash}p{0.39\columnwidth}
  r@{}}
\toprule
ID & Motion family & Condition & Count (\%) \\
\midrule
0 & Stop/Crawl        & $D_n<5$                                             & 12,044 (11.66) \\
1 & Cruise            & $|\bar\psi_n|\le4.5^\circ$, $|\Delta\bar s_n|\le\eta_s$ & 8,695 (8.42) \\
2 & Accelerating      & $|\bar\psi_n|\le4.5^\circ$, $\Delta\bar s_n>\eta_s$     & 17,339 (16.79) \\
3 & Decelerating      & $|\bar\psi_n|\le4.5^\circ$, $\Delta\bar s_n<-\eta_s$    & 24,501 (23.72) \\
4 & Left lane change  & $4.5^\circ<\bar\psi_n\le13^\circ$              & 9,330 (9.03) \\
5 & Left turn         & $\bar\psi_n>13^\circ$                           & 18,174 (17.60) \\
6 & Right lane change & $-13^\circ\le\bar\psi_n<-4.5^\circ$            & 5,422 (5.25) \\
7 & Right turn        & $\bar\psi_n<-13^\circ$                          & 7,783 (7.54) \\
\bottomrule
\end{tabular}
}
\caption{Rule-defined, trajectory-derived M-GMP maneuver families and their occupancy in
the NAVSIM \texttt{navtrain} data used for prior construction.}
\label{tab:app_mgmp_families}
\end{table}

\subsection{Soft Component Association and Training Routes}
After fitting, the component-normalized diff-space distance for each expert--component pair is
\begin{equation}
\begin{aligned}
d_{n,m}
&=\frac{1}{H_\tau}\sum_{h=1}^{H_\tau}\left[
\left\|
(\Delta\widetilde{\mathbf p}_{n,h}-\boldsymbol\mu^p_{m,h})
\oslash\overline{\boldsymbol\sigma}^p_{m,h}
\right\|_2^2\right.\\[-2pt]
&\hspace{68pt}\left.+
\frac{\operatorname{wrap}^2(\psi_{n,h}-\mu^\psi_{m,h})}
{(\overline{\sigma}^\psi_{m,h})^2}
\right],
\end{aligned}
\label{eq:app_component_distance}
\end{equation}
where $\oslash$ is element-wise division and
$\operatorname{wrap}(\cdot)\in(-\pi,\pi]$. Normalizing the component
similarities gives the soft maneuver supervision target
\begin{equation}
y_{n,m}^{\mathrm{man}}
 =\frac{\exp(-d_{n,m})}
 {\sum_{\ell=0}^{M-1}\exp(-d_{n,\ell})}.
\label{eq:app_soft_maneuver_target}
\end{equation}
Its highest-similarity component defines the discrete match used for routing:
\begin{equation}
m_n^\star
 =\arg\max_m y_{n,m}^{\mathrm{man}}
=\arg\min_m d_{n,m}.
\label{eq:app_matched_component}
\end{equation}
The rule label $m_n^{\mathrm{rule}}$ only constructs and names the components.
The distance-based component similarity target $y_n^{\mathrm{man}}$ supervises $\pi_n^{\mathrm{man}}$;
$m_n^\star$ defines the route used by the generation objective and the
Recall@$k$ target for the Maneuver-Mixture MeanFlow Field (M$^3$F). The rule
and matched labels serve distinct roles and are not assumed identical per
trajectory. Stage-2 candidates are generated by the current model using its
predicted $\pi_n^{\mathrm{man}}$ (Sec.~\ref{app:training_supervision});
deployment allocation uses the final model's prediction for the current scene.
Accordingly, the Recall@$k$ value in the main paper measures whether the
distance-matched fitted component $m_n^\star$ is contained in the predicted
top-$k$ set. It therefore quantifies scene-conditioned support coverage for the
operational intention representation; it does not measure classification accuracy against separate
human-labeled semantic maneuver annotations.

\paragraph{M$^3$F interval sampling.}
The distance-matched component $m_n^\star$ also determines the M$^3$F training
route. For each scene, we independently sample six source proposals with
$m_{n,i}=m_n^\star$; all six share one interval pair $(r,t)$. We sample this
pair from three regimes:
\begin{equation}
(r,t)=
\begin{cases}
(0,1), & \text{with probability }0.25,\\
(q,q), & \text{with probability }0.375,\\
(q_{(1)},q_{(2)}), & \text{with probability }0.375.
\end{cases}
\label{eq:app_interval_sampling}
\end{equation}
Here, $q=\operatorname{sigmoid}(\xi)$,
$\xi\sim\mathcal N(-0.4,1)$, and $q_1,q_2$ are independent draws, with
$q_{(1)}=\min(q_1,q_2)$ and $q_{(2)}=\max(q_1,q_2)$. The FP32 JVP uses the
boundary-field tangent and stop-gradient JVP correction defined in the main
paper.

\section{Verify: Shared-Future Evidence and Factorized Ranking}
\label{app:evidence_construction}

\paragraph{Notation bridge.}
The main paper uses $B_n$ as an abstract shared BEV-map interface. Here we
expose its implementation views: $F_n^{\mathrm{hi}}$ is the high-resolution
current map, $F_n^{\mathrm{lo}}$ is the low-resolution grid consumed by the
world model, and $B_n^{\mathrm{tok}}$ (defined explicitly in
Eq.~\eqref{eq:app_bev_tokens}) is its projected token view used by the
scene/agent decoder and M$^3$F. These are encoder views derived from the same current
observation, not additional sensor inputs.

\subsection{Future-Feature Targets}
Given $F_n^{\mathrm{lo}}$ and $c_n^{\mathrm{SN}}$, the world model predicts
absolute and residual features at 1--4 s. At 2 Hz, the detached targets use
offsets $+2,+4,+6,+8$; each is inverse-SE(2) warped from its local ego frame to
the current frame using recorded poses. Bilinear \texttt{grid\_sample} uses
zero padding, \texttt{align\_corners=True}, and an overlap mask for missing or
out-of-view cells. The first residual is relative to $F_n^{\mathrm{lo}}$ and
later residuals are between consecutive aligned targets. Future observations
remain confined to this stop-gradient path and are unavailable at deployment.

\subsection{Trajectory-Guided Evidence Queries}
For current evidence, $F_n^{\mathrm{hi}}$ is pooled to the future-field grid
and projected into reader space:
\begin{equation}
\overline F_n^{\mathrm{cur}}
=\operatorname{Conv}_{1\times1}
\!\left(\operatorname{Pool}(F_n^{\mathrm{hi}})\right).
\label{eq:app_current_grid}
\end{equation}
Let $(\mathbf p_{n,i,h},\psi_{n,i,h})$ be a candidate waypoint and
$\boldsymbol\delta_a$ a vehicle-frame footprint anchor from the corresponding
current or future query set. Its location in the current ego frame is
\begin{equation}
\mathbf q_{n,i,h,a}
=\mathbf p_{n,i,h}
+\mathbf R_{\mathrm{2D}}(\psi_{n,i,h})\boldsymbol\delta_a ,
\label{eq:app_footprint_query}
\end{equation}
where $\mathbf R_{\mathrm{2D}}(\psi)$ is the planar rotation matrix. The map
$\Pi:\mathbb R^2\rightarrow[-1,1]^2$ converts this location to normalized BEV
coordinates. Current evidence is obtained by bilinear sampling at the projected BEV coordinates,
\begin{equation}
\mathbf e_{n,i,h,a}^{\mathrm{cur}}
=\operatorname{Bil}
\!\left(\overline F_n^{\mathrm{cur}},
\Pi(\mathbf q_{n,i,h,a})\right).
\label{eq:app_current_probe}
\end{equation}
At waypoint $h\in\{2,4,6,8\}$, future evidence queries absolute/residual state
$j=h/2$ (1--4 s); out-of-support samples are masked to zero. The reader fuses
these path-aligned features with trajectory, ego-state, and route embeddings to
form $E_{n,i}$. Candidates therefore share one predicted future field while
retrieving path-specific evidence.

\subsection{Factor-Wise Metric Prediction and Ranking}
For a deployment-sized set of $N=64$ candidates, Metric Prism forms one
512-D score token per trajectory. A shared four-layer set-scoring stack jointly
updates the $N$ tokens, and token $i$ cross-attends to its associated
trajectory evidence $E_{n,i}$. Denote the resulting token by $q_{n,i}^{S}$.
Five factor-specific scalar heads, shared across candidate positions, operate
over the factor index set
\begin{equation}
\mathcal J_{\mathrm{PDM}}
 =\{\mathrm{NC},\mathrm{DAC},\mathrm{EP},
\mathrm{TTC},\mathrm{Comf}\}.
\label{eq:app_factor_index_set}
\end{equation}
For each $f\in\mathcal J_{\mathrm{PDM}}$, the corresponding head predicts
\begin{equation}
s_{n,i}^{f}
=\mathcal D_f(q_{n,i}^{S})\in(0,1).
\label{eq:app_factor_heads}
\end{equation}
Here $\mathcal D_f$ includes its terminal sigmoid. The heads are
factor-specific; the candidate representation remains shared across factor heads and
set-contextualized.

During Stage~2, NAVSIM v1 evaluator targets $\mathbf y_{n,i}\in[0,1]^5$
supervise factor predictions for the $N=64$ candidates in each training set
(Sec.~\ref{app:training_supervision}):
\begin{equation}
\mathcal L_{\mathrm{score}}
=\sum_{i=1}^{N}\sum_{f\in\mathcal J_{\mathrm{PDM}}}
\ell_f\!\left(s_{n,i}^{f},y_{n,i}^{f}\right).
\label{eq:app_factor_loss}
\end{equation}
The weighted binary cross-entropy (BCE) and logit regularization included in
$\ell_f$ are
\begin{equation}
\ell_f(s,y)=\frac{10}{N}\left[
\alpha_f w_f(y)\operatorname{BCE}(s,\widetilde y_f)
+\frac{10^{-5}}{5}\operatorname{logit}(s)^2
\right],
\label{eq:app_factor_loss_detail}
\end{equation}
where $\operatorname{BCE}(s,t)=-t\log s-(1-t)\log(1-s)$.
In the factor order of Eq.~\eqref{eq:app_factor_index_set},
$\boldsymbol\alpha=(5,5,5,5,2)$.
The smoothed target is
$\widetilde y_f=(1-\varepsilon_f)u_f(y)+\varepsilon_f/2$,
with $u_{\mathrm{NC}}(y)=\mathbf 1[y>0.5]$ and $u_f(y)=y$ for the other
factors; $\varepsilon_{\mathrm{NC}}=0.02$ and $\varepsilon_f=0.01$ otherwise.
NC sample weights are $1.7$, $3.1$, and $1.0$ for raw labels $0$, $0.5$,
and $1$, respectively. The remaining weights are
$w_{\mathrm{DAC}}(y)=1.8-0.8y$,
$w_{\mathrm{EP}}(y)=1$,
$w_{\mathrm{TTC}}(y)=1.75-0.75y$, and
$w_{\mathrm{Comf}}(y)=4.5-3.5y$.
The factor weights are summed, and the loss is averaged over candidates and
the mini-batch. The leading factor $10$ is the score-loss scale within
$\mathcal L_{\mathrm{verify}}$.

At inference, the five predictions define a scalar ranking score through the
fixed log-PDM composition:
\begin{equation}
\begin{aligned}
S_{n,i}
={}&\log(s_{n,i}^{\mathrm{NC}}+\epsilon)
+\log(s_{n,i}^{\mathrm{DAC}}+\epsilon)\\
&+\log\!\left(
5s_{n,i}^{\mathrm{EP}}
+5s_{n,i}^{\mathrm{TTC}}
+2s_{n,i}^{\mathrm{Comf}}
+\epsilon\right),
\end{aligned}
\label{eq:app_log_pdm}
\end{equation}
with $\epsilon=10^{-6}$. The conventional $1/12$ normalization is omitted
because its candidate-independent log offset preserves the ranking.

The same trajectory evidence conditions ECFR, whose supervision is defined next.

\FloatBarrier
\section{Improve: Candidate Supervision and Frenet Refinement}
\label{app:training_supervision}

\subsection{Stage-2 Candidates and Supervision}
During Stage~2, all inference-time modules are jointly optimized.
The current model generates a fresh $N=64$ candidate set for each training
scene. Metric Prism and ECFR learn from this same set, using evidence extracted
by the trainable current/future BEV branches. For each \texttt{navtrain} scene,
let $\mathcal K_n=\operatorname{Top}_3(\pi_n^{\mathrm{man}})$ and renormalize
the selected component weights within this set:
\begin{equation}
\widetilde\pi_{n,m}
 =\frac{\pi_{n,m}^{\mathrm{man}}}
{\sum_{q\in\mathcal K_n}\pi_{n,q}^{\mathrm{man}}},
 \qquad m\in\mathcal K_n.
\label{eq:app_training_top3_weights}
\end{equation}
Let $\widetilde{\boldsymbol\pi}_n
=(\widetilde\pi_{n,m})_{m\in\mathcal K_n}$. The route indices are sampled as
\begin{equation}
m_{n,i}
 \overset{\mathrm{i.i.d.}}{\sim}
\operatorname{Cat}(\widetilde{\boldsymbol\pi}_n),
\qquad i=1,\ldots,N,\quad N=64.
\label{eq:app_training_component_draw}
\end{equation}
Each draw selects an M-GMP source and its M$^3$F expert. One-step transport
produces $\mathcal T_n^{\mathrm{base}}=\{\tau_{n,i}\}_{i=1}^{64}$, with each
candidate retaining its generating index $m_{n,i}$.
The NAVSIM v1 evaluator supplies the corresponding five-factor labels.
These labels supervise Metric Prism and determine ECFR reference trajectories
within the same set.

\subsection{Evaluator Labels and Refinement References}
For each candidate $\tau_{n,i}\in\mathcal T_n^{\mathrm{base}}$, the evaluator
returns the normalized factor vector
\begin{equation}
\mathbf y_{n,i}
=(y^{\mathrm{NC}},y^{\mathrm{DAC}},y^{\mathrm{EP}},
y^{\mathrm{TTC}},y^{\mathrm{Comf}})_{n,i}\in[0,1]^5 .
\label{eq:app_evaluator_factor_vector}
\end{equation}
The raw factor labels define the reference-selection score
\begin{equation}
P_{\mathrm{eval}}(\tau_{n,i})
:=y_{n,i}^{\mathrm{NC}}y_{n,i}^{\mathrm{DAC}}\times
\frac{5y_{n,i}^{\mathrm{EP}}+5y_{n,i}^{\mathrm{TTC}}
+2y_{n,i}^{\mathrm{Comf}}}{12}.
\label{eq:app_evaluator_pdm}
\end{equation}
For each base candidate $i$, the same-maneuver maximizer in the current
64-candidate set provides its reference trajectory for Frenet refinement:
\begin{equation}
\tau_{n,i}^\dagger
=\operatorname*{arg\,max}_{\substack{
\tau_{n,j}\in\mathcal T_n^{\mathrm{base}}\\
m_{n,j}=m_{n,i}}}
P_{\mathrm{eval}}(\tau_{n,j}).
\label{eq:app_refinement_reference}
\end{equation}
Reference selection considers candidates with finite trajectories and scores.
The base itself is eligible, so a best-scoring base may have a zero residual
target. Invalid base--reference pairs use zero residual targets and weights.

\subsection{Base-Local Frenet Targets}
Write the selected reference as
$\tau_{n,i}^\dagger
=\{(\mathbf p^\dagger_{n,i,h},\psi^\dagger_{n,i,h})\}_{h=1}^{H_\tau}$.
For base waypoint $(\mathbf p_{n,i,h},\psi_{n,i,h})$, we compute a centered
path difference at interior waypoints and a one-sided difference at each
endpoint, then normalize it to obtain $\mathbf t_{n,i,h}$. At a degenerate
tangent, we use
$(\cos\psi_{n,i,h},\sin\psi_{n,i,h})$; the corresponding normal is
$\mathbf n_{n,i,h}=(-t^y_{n,i,h},t^x_{n,i,h})$. Projecting the reference--base
position difference onto the tangent gives the longitudinal target
\begin{equation}
\overline\Delta^s_{n,i,h}
 =(\mathbf p^\dagger_{n,i,h}-\mathbf p_{n,i,h})^\mathsf T
 \mathbf t_{n,i,h}.
\label{eq:app_frenet_longitudinal_target}
\end{equation}
The projection onto the corresponding normal gives the per-waypoint lateral displacement target
\begin{equation}
\overline\Delta^l_{n,i,h}
 =(\mathbf p^\dagger_{n,i,h}-\mathbf p_{n,i,h})^\mathsf T
 \mathbf n_{n,i,h}.
\label{eq:app_frenet_lateral_target}
\end{equation}
Finally, the wrapped reference--base heading difference defines the angular target
\begin{equation}
\overline\Delta^\psi_{n,i,h}
 =\operatorname{wrap}(\psi^\dagger_{n,i,h}-\psi_{n,i,h}).
\label{eq:app_frenet_heading_target}
\end{equation}

The inverse map $\mathcal U_{\mathrm F}$ applies
$\widetilde{\mathbf p}=\mathbf p+\Delta^s\mathbf t+\Delta^l\mathbf n$ and
$\widetilde\psi=\operatorname{wrap}(\psi+\Delta^\psi)$ pointwise.
Stacking $(\overline\Delta^s_{n,i,h},\overline\Delta^l_{n,i,h},
\overline\Delta^\psi_{n,i,h})$ over $h=1,\ldots,H_\tau$ gives the main-paper
target $\overline\Delta^{\mathrm F}_{n,i}
=\mathcal R_{\mathrm F}(\tau_{n,i},\tau_{n,i}^\dagger)
\in\mathbb R^{H_\tau\times3}$.
The residual loss $\ell_{\mathrm F}$ is Smooth-$L_1$ with unit transition,
averaged over all $H_\tau$ waypoints and the longitudinal, lateral, and heading
coordinates. The validity mask $\omega_{n,i}$ is $1$ for a finite
base--reference pair and $0$ otherwise. The main-paper
$\mathcal L_{\mathrm{ref}}$ averages over valid supervised candidates and
applies $\alpha_{\mathrm{ref}}=5$; the same normalization is used across the
mini-batch. Reference selection and residual targets are detached.

Table~\ref{tab:app_reproducibility} consolidates the module interfaces and the
two-stage training protocol before Sec.~\ref{app:reproducibility} expands the
input, optimization, and deployment details.

\begin{table*}[!t]
\centering
{\small
\renewcommand{\arraystretch}{1.03}
\setlength{\tabcolsep}{4.0pt}
\begin{tabular*}{\textwidth}{@{\extracolsep{\fill}}
  >{\centering\arraybackslash}m{0.10\textwidth}
  >{\centering\arraybackslash}m{0.15\textwidth}
  >{\raggedright\arraybackslash}m{0.32\textwidth}
  >{\raggedright\arraybackslash}m{0.26\textwidth}
  >{\centering\arraybackslash}m{0.10\textwidth}@{}}
\multicolumn{5}{@{}l}{\textbf{(a)\quad Module architecture}} \\[-1pt]
\toprule
Lifecycle & Module & \multicolumn{1}{c}{Structure}
& \multicolumn{1}{c}{Config. $(d,h,d_{\mathrm{ff}})$} & Output \\
\midrule
\multirow{2}{*}{\textbf{Shared}}
& Perception encoder
& ResNet-34; 4-stage TransFuser; 4 GPT/stage
& $h=8$; $d_{\mathrm{ff}}=4d$; dropout 0.1
& $F_n^{\mathrm{hi}},F_n^{\mathrm{lo}}$ \\
& Scene/agent decoder
& Two-layer Transformer decoder
& $(256,8,1024)$; dropout 0
& -- \\
\midrule
\textbf{Hypothesize}
& M$^3$F
& 6 DiT--MoE; 8 FFN experts/block
& $(256,4,1024)$; attn./FFN 0/0.1
& $\mathbb R^{8\times4}$ \\
\midrule
\multirow{3}{*}{\textbf{Verify}}
& World model
& Recurrent predictor; causal refinement
& --
& 4 abs. / 4 res. \\
& Evidence reader
& Trajectory-guided BEV sampling; SlotPool
& Token 512; $h=8$; dropout 0.1
& $E_{n,i}$ \\
& Metric Prism
& 4-layer set scorer; 5 factor heads
& $(512,1,1024)$; proj./path 0.1/0.2
& 5 factors \\
\midrule
\textbf{Improve}
& ECFR
& 2-layer Frenet Transformer
& $(512,8,1024)$; dropout 0.1
& $\mathbb R^{8\times3}$ \\
\bottomrule
\end{tabular*}
\par\smallskip
\begin{tabular*}{\textwidth}{@{\extracolsep{\fill}}
  >{\raggedright\arraybackslash}p{0.20\textwidth}
  >{\raggedright\arraybackslash}p{0.37\textwidth}
  >{\raggedright\arraybackslash}p{0.37\textwidth}@{}}
\multicolumn{3}{@{}l}{\textbf{(b)\quad Two-stage optimization}} \\[-1pt]
\toprule
Setting & Stage 1 & Stage 2 \\
\midrule
Trainable modules
& Perception, auxiliary heads, and M$^3$F
& All inference-time modules under one Stage-2 objective \\
Training candidate source
& --
& Current encoder, maneuver predictor, and M$^3$F; 64 candidates per scene \\
Supervision
& $\mathcal L_{\mathrm{aux}}+\mathcal L_{\mathrm{M^3F}}$ only
& Detached future field (1--4 s); NAVSIM v1 factor labels and Frenet targets
  from the same 64-candidate set \\
Loss weights
& 1 / 4 / -- / --
& 1 / 4 / 2 / 2 \\
\addlinespace[1.5pt]
Base learning rate
& $4{\times}10^{-4}$
& $2{\times}10^{-4}$ \\
Perception/M$^3$F LR scale
& 1.0
& 0.2 \\
\addlinespace[1.5pt]
Optimizer / weight decay & AdamW / 0.1 & AdamW / 0.1 \\
Training data & NAVSIM \texttt{navtrain} & NAVSIM \texttt{navtrain} \\
Iterations / epoch & 2,660 & 2,660 \\
Training epochs & 60 & 10 \\
Total optimizer updates & 159,600 & 26,600 \\
Global batch / precision & 32 / FP32 & 32 / FP32 \\
Clipping / hardware
& Global norm 1.0 / 4$\times$ RTX 4090
& Global norm 1.0 / 4$\times$ RTX 4090 \\
\bottomrule
\end{tabular*}
}
\caption{Reproducibility overview. \textbf{(a)} Module architecture; outputs
omit the batch dimension (interfaces:
Sec.~\ref{sec:app_multires_interface}). \textbf{(b)} Two-stage
optimization; loss weights follow auxiliary / generation / verification /
refinement. The score and residual losses include their internal scales
of 10 and 5, respectively. Stage~2 initializes from Stage~1 and jointly optimizes
all inference-time modules; the current model generates candidates for evaluator labeling.}
\label{tab:app_reproducibility}
\label{tab:app_architecture}
\label{tab:app_training}
\end{table*}

\section{Reproducibility: Interfaces, Optimization, and Deployment}
\label{app:reproducibility}
\label{app:architecture}
\label{app:training_protocol}

Table~\ref{tab:app_reproducibility}(a) summarizes the lifecycle modules and
their interfaces, while panel (b) summarizes the two-stage optimization. The
following subsections unpack the inputs, shared representation, training protocol, and
deployment lifecycle for both TF and LTF.

\subsection{Input Construction}
The LiDAR-assisted TransFuser (TF) and camera-only Latent TransFuser (LTF) use
current left, front, and right views. All are cropped by 28 pixels vertically;
the lateral views are also cropped by 416 pixels per horizontal side before
concatenation and resizing to $256\times1024$. TF rasterizes one LiDAR scan over
$[-32,32]^2$ m at $0.25$ m per pixel, retains $z>0.2$ m, and clips then divides
pixel counts by five. LTF replaces LiDAR with a learnable latent template
$\mathbf Z_{\mathrm{lat}}\in\mathbb R^{1\times4\times256\times256}$, where the
leading singleton dimension indexes one shared template, the four channels are
latent features, and the $256\times256$ grid spans the BEV. At runtime, the
template is broadcast along the batch dimension, yielding the actual encoder
input shape $[B,4,256,256]$. Accordingly, the LTF ResNet stem uses a
convolution-weight tensor of shape $[64,4,7,7]$.
Both receive a 4-D command code, 2-D velocity, and 2-D acceleration.

\begin{table*}[!t]
\centering
{\small
\begin{minipage}[t]{0.31\textwidth}
\centering
\textbf{(a)\quad NAVSIM v1 navtest}\par\vspace{2pt}
\setlength{\tabcolsep}{2.0pt}
\begin{tabular}{@{}lcc@{}}
\toprule
Metric & LTF & TF \\
\midrule
NC    & $98.49{\pm}0.02$ & $98.42{\pm}0.02$ \\
DAC   & $97.40{\pm}0.03$ & $98.01{\pm}0.03$ \\
TTC   & $95.13{\pm}0.05$ & $95.19{\pm}0.05$ \\
Comf. & $99.99{\pm}0.00$ & $99.99{\pm}0.01$ \\
EP    & $84.87{\pm}0.04$ & $85.36{\pm}0.04$ \\
PDMS  & $89.98{\pm}0.06$ & $90.58{\pm}0.05$ \\
\bottomrule
\end{tabular}
\end{minipage}
\hfill
\begin{minipage}[t]{0.31\textwidth}
\centering
\textbf{(b)\quad NAVSIM v2 navtest}\par\vspace{2pt}
\setlength{\tabcolsep}{2.0pt}
\begin{tabular}{@{}lcc@{}}
\toprule
Metric & LTF & TF \\
\midrule
NC     & $98.61{\pm}0.02$ & $98.38{\pm}0.03$ \\
DAC    & $98.13{\pm}0.05$ & $97.94{\pm}0.03$ \\
DDC    & $99.25{\pm}0.02$ & $99.41{\pm}0.02$ \\
TLC    & $99.80{\pm}0.01$ & $99.76{\pm}0.00$ \\
EP     & $87.27{\pm}0.01$ & $88.75{\pm}0.01$ \\
TTC    & $97.96{\pm}0.03$ & $97.75{\pm}0.02$ \\
LK     & $96.13{\pm}0.03$ & $96.86{\pm}0.06$ \\
HC     & $98.26{\pm}0.02$ & $98.32{\pm}0.02$ \\
EC     & $85.17{\pm}0.14$ & $86.54{\pm}0.07$ \\
EPDMS  & $89.70{\pm}0.06$ & $90.14{\pm}0.04$ \\
\bottomrule
\end{tabular}
\end{minipage}
\hfill
\begin{minipage}[t]{0.34\textwidth}
\centering
\textbf{(c)\quad WorldEngine}\par\vspace{2pt}
\setlength{\tabcolsep}{2.4pt}
\begin{tabular}{@{}llc@{}}
\toprule
Traffic & Metric & ManeuverFlow-LTF \\
\midrule
Log & SR          & $87.48{\pm}0.92$ \\
Log & EP          & $60.68{\pm}0.91$ \\
Log & PDMS$^\ast$ & $75.19{\pm}0.85$ \\
IDM & SR          & $87.92{\pm}0.87$ \\
IDM & EP          & $60.89{\pm}0.81$ \\
IDM & PDMS$^\ast$ & $74.58{\pm}0.64$ \\
\bottomrule
\end{tabular}
\end{minipage}
}
\caption{Main-result variation, reported as mean $\pm$ sample standard
deviation across three independently trained checkpoints, each averaged over
five evaluations. WorldEngine uses log replay (Log) or reactive IDM traffic.}
\label{tab:app_main_statistics}
\end{table*}

\begin{table*}[!t]
\centering
\begin{minipage}[t]{0.48\textwidth}
\vspace{0pt}
\centering
\small
\setlength{\tabcolsep}{8pt}
\begin{tabular}{@{}c cc@{}}
\toprule
Top-$\kappa$ & Exact oracle & Within 1 pt \\
\midrule
1  & 40.14 & 75.28 \\
3  & 45.61 & 79.57 \\
5  & 48.14 & 81.16 \\
10 & 52.69 & 83.18 \\
\bottomrule
\end{tabular}
\caption{Within-pool Top-$\kappa$ retrieval (\%) of Metric Prism on the unrefined
Base64 pool. ``Within 1 pt'' allows one evaluator PDMS point relative to that
pool's oracle.}
\label{tab:app_score_ranking}
\end{minipage}
\hfill
\begin{minipage}[t]{0.48\textwidth}
\vspace{0pt}
\centering
\setlength{\tabcolsep}{2.2pt}
\begin{tabular}{@{}rcccc@{}}
\toprule
$N$ & Best & Top-3 & Top-6 & Top-12 \\
\midrule
8   & 92.20 & 90.90 & 89.18 & -- \\
16  & 92.92 & 91.98 & 91.00 & 89.19 \\
32  & 93.57 & 92.81 & 92.06 & 91.05 \\
64  & 94.25 & 93.54 & 92.92 & 92.14 \\
128 & 94.61 & 94.03 & 93.55 & 92.93 \\
256 & 94.85 & 94.47 & 94.08 & 93.58 \\
512 & 95.02 & 94.79 & 94.47 & 94.08 \\
\bottomrule
\end{tabular}
\caption{Frozen-generator $\operatorname{OracleTop}_{\kappa}(N)$ diagnostic
on NAVSIM v1 \texttt{navtest}, using nested prefixes of one fixed per-scene
A512 draw on the 0--100 PDMS scale.}
\label{tab:app_oracle_topk}
\end{minipage}
\end{table*}

\subsection{Multi-Resolution Perception Interface}
\label{sec:app_multires_interface}
The shared encoder exposes
$F_n^{\mathrm{hi}}\in\mathbb R^{b\times64\times64\times64}$ and a
low-resolution dynamic map $F_n^{\mathrm{lo}}$. The former supports auxiliary
BEV semantics and pooled current evidence; a $1\times1$ projection maps the
latter to
\begin{equation}
B_n^{\mathrm{tok}}
:=\operatorname{Flat}\!\left(
\operatorname{Conv}_{1\times1}(F_n^{\mathrm{lo}})
\right)
\in\mathbb R^{b\times64\times256}.
\label{eq:app_bev_tokens}
\end{equation}
Here, $b$ is the batch size. $B_n^{\mathrm{tok}}$ conditions the scene/agent
decoder and M$^3$F;
$F_n^{\mathrm{lo}}$ drives future prediction, and the reader projects pooled
current and predicted-future features into its query space. The main paper uses $B_n$ as a
compact symbol for this multi-resolution BEV interface rather than for the projected token sequence
$B_n^{\mathrm{tok}}$ alone.

\subsection{Two-Stage Optimization}
Table~\ref{tab:app_training}(b) summarizes the two stages. Stage~2 initializes
the trainable model from Stage~1 and uses the 64-candidate supervision
defined in Sec.~\ref{app:training_supervision}. Perception and M$^3$F continue
to receive the auxiliary and generation objectives, while the shared BEV
features support future prediction, scoring, and refinement. All inference-time
modules are jointly optimized. Candidate construction, evaluator labels,
reference selection, Frenet targets, and future-feature targets are detached from gradient computation.

\subsection{Deployment Lifecycle}
At each planning step, the final model generates 64 base candidates using
Top-3 largest-remainder allocation and predicts one shared future BEV field
from the current observation. Metric Prism first ranks the bases using their
trajectory evidence. ECFR refines 16 selected bases and replaces their slots,
retaining the other 48. All 64 final hypotheses re-query the same BEV field
and are jointly re-scored by Metric Prism. The candidate with the highest
predicted log-PDM score is returned as the final plan.

Table~\ref{tab:app_main_statistics} reports the run-to-run variation across independently trained checkpoints analyzed in
Sec.~\ref{app:result_reliability}.

\begin{figure*}[!t]
\centering
\includegraphics[width=0.98\textwidth]%
  {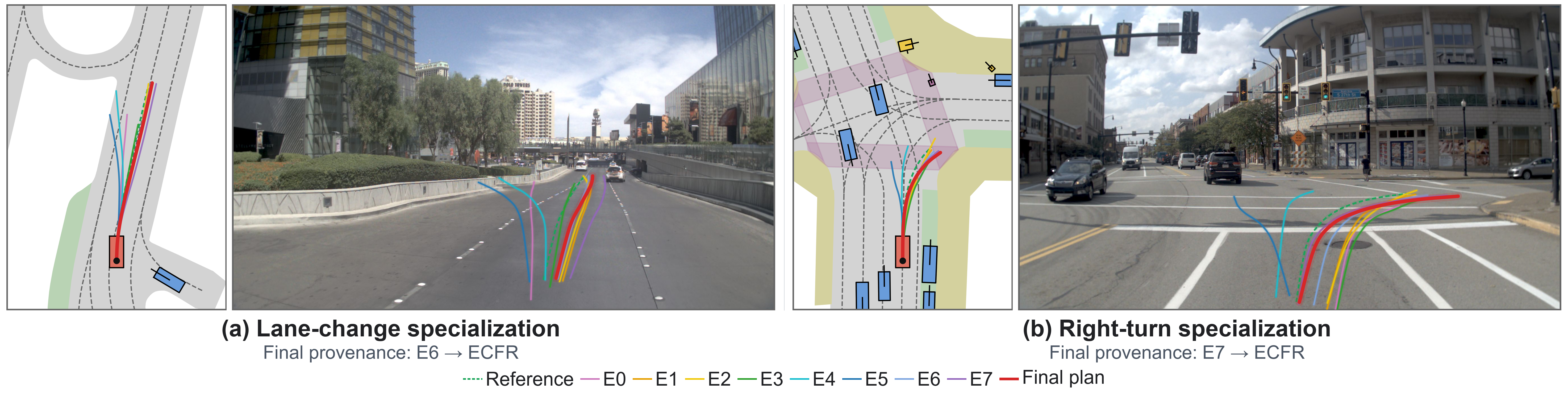}
\caption{\textbf{Controlled expert sweep and final-plan provenance.}
\textbf{(a,b)} Paired BEV/front views of two NAVSIM scenes. Thin E0--E7
trajectories are controlled responses sharing a fixed M-GMP source state,
scene context, and transport interval; the source is omitted, dashed green is
the logged reference, and thick red is the final trajectory $\tau_n^\star$.
The final plan comes from the full $N=64$ base/refined pool after ECFR,
evidence re-query, and Metric Prism re-scoring rather than from the displayed
sweep, with provenance E6$\rightarrow$ECFR in (a) and E7$\rightarrow$ECFR in
(b). These examples expose route-conditioned response separation and final-plan
provenance; dataset-level support is supplied by the quantitative
component-support and matched-ablation results.}
\label{fig:app_m3f_expert_specialization}
\end{figure*}

\section{Statistical Reliability and Scoring Diagnostics}
\label{app:result_reliability}

\subsection{Main-Result Variation}
Across three independently trained checkpoints, the largest observed standard
deviation is 0.14 on NAVSIM and 0.92 in WorldEngine. These values describe
run-to-run dispersion under the stated protocol; with three training runs,
they should not be interpreted as formal confidence intervals.
Table~\ref{tab:app_main_statistics} provides the corresponding dispersion for
every ManeuverFlow row in the three
main-result tables. It does not provide matched-seed uncertainty for the
component ablations in the main paper.

\subsection{Within-Pool Top-$\kappa$ Retrieval of Metric Prism}
We evaluate Metric Prism's preliminary ordering on the same 12,146 NAVSIM v1
\texttt{navtest} scene tokens used by the frozen-pool analysis. For every
trajectory in the unrefined Base64 pool, Metric Prism predicts the five PDM
factors and applies Eq.~\eqref{eq:app_log_pdm}. Here Base64 is shorthand for
the unrefined 64-candidate pool used by this retrieval diagnostic; all
percentages are relative to its own evaluator oracle. The main paper's $k$
denotes the number of routed components, whereas $\kappa$ here denotes ranking
depth. The diagnostic asks whether
the first $\kappa$ positions contain the evaluator-best trajectory in that
particular pool, either exactly or within one PDMS point of its oracle score.

As Table~\ref{tab:app_score_ranking} shows, at rank one, Metric Prism retrieves
a trajectory within one PDMS point of the
Base64 oracle in 75.28\% of scenes and misses this tolerance in 24.72\%; by
rank ten, coverage reaches 83.18\%. These are pool-relative retrieval
statistics, not a budget-matched cross-method scorer comparison. This
diagnostic characterizes the learned scorer on the evaluated unrefined Base64
pool; it does not isolate the future branch or compare Metric Prism with
another scorer.

\subsection{Fixed Selector Transfer to NAVSIM v2}
Table~\ref{tab:app_main_statistics}(b) evaluates the complete v1-trained
checkpoints under NAVSIM v2 while retaining the same five factor heads and
log-PDM selector. The v2-only factors contribute to EPDMS evaluation but are
not added to the selector. The resulting EPDMS therefore provides an empirical
feasibility check for transferring the fixed selector to the extended
evaluator, rather than a claim that v1 PDMS and v2 EPDMS are directly
comparable across evaluator versions.

\subsection{Frozen-Pool Headroom Diagnostic}
\label{app:ecfr_sampling_budget}
For each scene in \texttt{navtest}, we construct a separate post-hoc analysis
draw of 512 candidates, denoted A512, using the Top-3 categorical sampling
distribution in Eq.~\eqref{eq:app_training_component_draw}, extended to 512
draws for this diagnostic. This analysis uses frozen checkpoints after
training and model selection. Across the 12,146 scenes in
$\mathcal D=\texttt{navtest}$, the
frozen M$^3$F generator produces one fixed, ordered A512 draw per scene. Its
nested prefixes
$\mathcal{T}_{N}^{\mathrm{ana}}(\nu)$ use
$N\in\{8,16,32,64,128,256,512\}$. Let
$S_\nu^{\mathrm{eval}}(\tau)$ denote evaluator PDMS on the 0--100 scale,
computed with the same factor aggregation as Eq.~\eqref{eq:app_evaluator_pdm}.
For $\kappa\le N$, we compute
\begin{equation}
  \operatorname{OracleTop}_\kappa(N)
  = \frac{1}{|\mathcal{D}|}
    \sum_{\nu \in \mathcal{D}}
    \frac{1}{\kappa}\sum_{a=1}^{\kappa} S_{\nu,(a)}^{\mathrm{eval}}.
  \label{eq:app_oracle_topk}
\end{equation}
Here, $S_{\nu,(a)}^{\mathrm{eval}}$ is the $a$-th largest prefix score;
$\kappa=1$ is best-of-$N$, whereas $\kappa>1$ averages the top-ranked region.
Nested prefixes measure headroom within the same draw, not scaling under
independent resampling or generator changes.

Table~\ref{tab:app_oracle_topk} quantifies candidate-quality headroom in a
frozen pool. It is
not an end-to-end budget sweep, an ECFR recovery measurement, or deployed
performance. Although the main-paper plot overlays the 94.8 \mbox{HumanAgent}
reference, its comparison with the A512 best-of-512 value 95.02 is an unpaired
aggregate order statistic; it establishes neither per-scene nor human-level
superiority.

\section{Qualitative Evidence and Claim Boundaries}
\label{app:qualitative_comparison}

\subsection{Controlled Expert Responses and Final-Plan Provenance}
Figure~\ref{fig:app_m3f_expert_specialization} holds the M-GMP source state,
scene context, and transport interval fixed while varying the routed expert. It
therefore provides a controlled mechanism check of the route-conditioned
response separation underlying intention-specific generation, and also traces
the final-plan provenance in two selected scenes. The figure complements rather
than substitutes for the dataset-level target-component Recall@$k$ and matched
prior/generator controls reported in the main paper's quantitative evaluation.

The right-lane-change example shows lateral and terminal-position variation, while
the right-turn example shows variation in turn onset, radius, and exit
alignment. The panel-title shorthand ``specialization'' denotes only this
operational, route-conditioned response separation under matched source and
context. The panels visualize the mechanism and are not themselves the deployed
candidate pool or a separate semantic labeling of the experts.

\subsection{Descriptive, Non-Budget-Matched Candidate Gallery}
\begin{figure*}[p]
\centering
\includegraphics[width=0.94\textwidth]{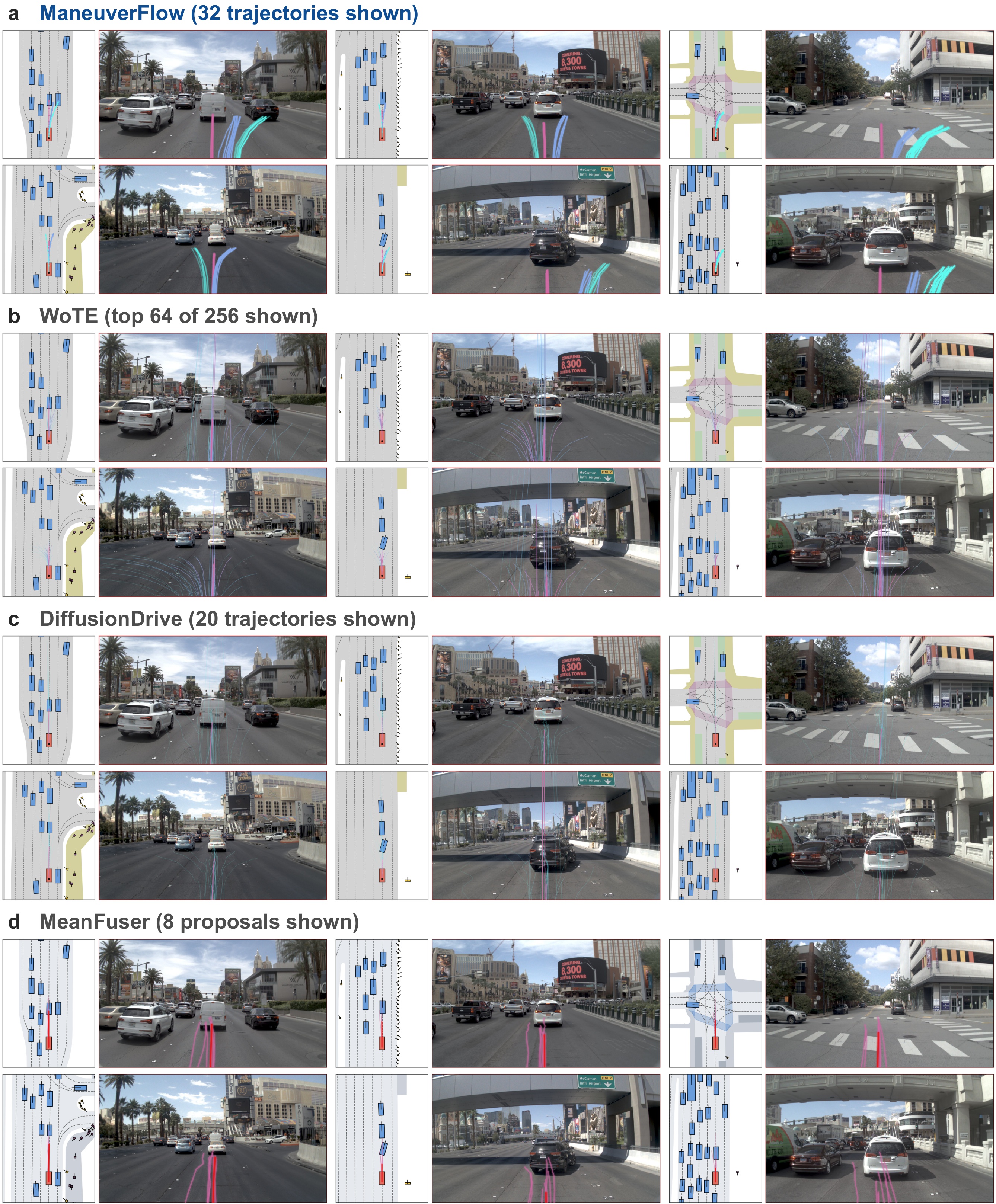}
\caption{Descriptive, non-budget-matched gallery on six shared scenes with
paired BEV/front views. (a) ManeuverFlow: a qualitative $N=32$ M$^3$F draw,
routed $16/8/8$ and colored by component; reported quantitative and deployment
results use $N=64$. (b) WoTE: the top 64 of 256 predictions by final reward.
(c) DiffusionDrive: all 20 candidates ordered by logits. (d) MeanFuser: eight
native proposal heads plus the fused output. Counts and selection are
method-specific; no quantitative comparison is implied.}
\label{fig:app_qualitative_comparison}
\end{figure*}

Figure~\ref{fig:app_qualitative_comparison} compares candidate organization on
the same six scenes using each method's display convention. Candidate counts,
ranking or filtering, and fused-output presentation differ across methods, so
the gallery is descriptive rather than budget-matched; no quantitative metric
is derived from it, and no random-sampling claim is made for the displayed
per-method candidate sets.

\newpage
\subsection{Interpretation and Evidence Scope}
\label{sec:app_limitations}

\paragraph{Interpretation scope.}
The paper operationalizes each intention as a trajectory-derived motion
hypothesis whose displacement, turning, or speed-trend tendency directly affects
planning behavior; it does not claim a separate map-semantic annotation task.
The quantitative claims concern scene-conditioned component support and
downstream planning gains, while the prior views and controlled expert sweeps
expose source organization and routed response separation. Top-three routing may
still miss rare or compositional behavior, and the qualitative examples and
learned PDM factors establish neither cross-method superiority nor safety. The
candidate-independent future field also does not model interactive responses to
alternative ego plans.

\paragraph{Experimental scope.}
The cumulative ablations evaluate the Hypothesize--Verify--Improve stages.
The Dense-XL control matches total generator parameter count, and Proposal FPS
measures perception through M$^3$F. Table~\ref{tab:app_oracle_topk} reports
post-hoc candidate-quality headroom within fixed analysis draws.
Table~\ref{tab:app_main_statistics} reports sample standard deviations across
independently trained checkpoints under the stated training and evaluation protocol.

\paragraph{Transfer scope.}
NAVSIM v2 applies the v1-trained checkpoints, five-factor heads, and selector
without v2-specific training or tuning, while WorldEngine reconstructs driving scenarios from
\texttt{navtest} logs. These protocols test fixed-selector transfer and
closed-loop replanning, not independent-domain or real-vehicle generalization.

\clearpage

\end{document}